\documentclass[lettersize,journal]{IEEEtran}
\usepackage{amsmath,amsfonts}
\usepackage{algorithmic}
\usepackage{algorithm}
\usepackage{array}
\usepackage[caption=false,font=normalsize,labelfont=sf,textfont=sf]{subfig}
\usepackage{textcomp}
\usepackage{stfloats}
\usepackage{url}
\usepackage{verbatim}
\usepackage{graphicx}
\usepackage{cite}
\usepackage{breqn}
\usepackage{makecell}
\begin{document}

\title{Parameter Optimization of Optical Six-Axis Force/Torque Sensor for Legged Robots}

\author{Hyun-Bin Kim, Byeong-Il Ham, Keun-Ha Choi, and Kyung-Soo Kim,~\IEEEmembership{Member,~IEEE,}
\thanks{Manuscript created September, 2024; This work was developed by the MSC (Mechatronics, Systems and Control) lab in the KAIST(Korea Advanced Institute of Science and Technology which is in the Daehak-Ro 291, Daejeon, South Korea(e-mail: youfree22@kaist.ac.kr; byeongil\_ham@kaist.ac.kr; choiha99@kaist.ac.kr; kyungsookim@kaist.ac.kr).(Corresponding author: Kyung-Soo Kim).}}
\markboth{ arXiv,
JANUARY~2025}%
{Shell \MakeLowercase{\textit{et al.}}: A Sample Article Using IEEEtran.cls for IEEE Journals}


\maketitle

\begin{abstract}
This paper introduces a novel six-axis force/torque sensor tailored for compact and lightweight legged robots. Unlike traditional strain gauge-based sensors, the proposed non-contact design employs photocouplers, enhancing resistance to physical impacts and reducing damage risk. This approach simplifies manufacturing, lowers costs, and meets the demands of legged robots by combining small size, light weight, and a wide force measurement range.
A methodology for optimizing sensor parameters is also presented, focusing on maximizing sensitivity and minimizing error. Precise modeling and analysis of objective functions enabled the derivation of optimal design parameters. The sensor's performance was validated through extensive testing and integration into quadruped robots, demonstrating alignment with theoretical modeling.
The sensor’s precise measurement capabilities make it suitable for diverse robotic environments, particularly in analyzing interactions between robot feet and the ground. This innovation addresses existing sensor limitations while contributing to advancements in robotics and sensor technology, paving the way for future applications in robotic systems.
\end{abstract}

\sloppy
\section{Introduction}
 \IEEEPARstart{R}{ecent} advancements in robotics have led to a growing interest in legged robots, exemplified by platforms such as \textit{ANYmal}, \textit{Mini Cheetah}, \textit{Hound}, and \textit{Raibo}~\cite{di2018dynamic, katz2019mini,hutter2016anymal,shin2022design,choi2023learning}. Force/torque sensors play a crucial role in these systems, enabling precise ground reaction force (GRF) measurement, which is essential for stability and control. Unlike manipulators, legged robots experience high impact forces often exceeding three to five times the nominal GRF due to foot-ground impacts. This repeated exposure to large forces often leads to sensor failure, making durability a key challenge.

To mitigate this issue, conventional sensors increase their sensing range to withstand high impact forces. However, this comes at the cost of reduced sensitivity, creating a fundamental trade-off between durability and measurement accuracy. Existing solutions, such as strain-gauge-based sensors, also suffer from high cost, large size, and the need for external signal processing, making them impractical for compact and lightweight robotic applications.

Force/torque sensors play a crucial role in robotics. In robotic arms, they are widely used in collaborative robots, while in humanoid robots~\cite{kim2019multi}, they measure ground reaction forces at the ankle for control purposes. In robotic hands, such as prosthetic devices, they facilitate force control through tension, and they are also extensively utilized in medical robotics~\cite{kim2017surgical,cho2022msc,jeong2018design}. Previous research on force/torque sensors has explored various sensing methods, including strain gauges~\cite{chen2025design}, capacitive methods~\cite{kim2016novel}, piezoresistive sensors~\cite{epstein2020bi}, magnetic sensing~\cite{ananthanarayanan2012compact}, and Fiber Bragg gratings~\cite{xiong2020six}. Among these, strain gauge-based sensors, capacitive methods, and optical methods have successfully achieved commercialization.

This paper introduces a novel non-contact six-axis force/torque sensor based on optical photo-couplers~\cite{kim2024compact,kim2023compact}, designed to address these challenges(high impact, sensitivity, durability, cost, large size and external signal processing). The proposed sensor significantly reduces impact-induced damage risks while maintaining high sensitivity. To overcome the trade-off between sensing range and sensitivity, an optimization-based approach was employed. Various objective functions were evaluated using finite element method (FEM) simulations to derive an optimal sensor design that balances robustness and accuracy.

The fabricated sensor was integrated into a quadruped robot and compared against a commercial reference sensor in real-world experiments. The results demonstrate that, while the commercial sensor exhibited significant offset drift due to repeated impacts, the proposed sensor maintained stable performance with minimal offset. Additionally, with only four components, the proposed sensor offers a cost-effective solution, with a prototype cost below \$300, significantly lower than the \$2,000-\$10,000 range of existing commercial sensors~\cite{robotous_website,ati_website}. These findings highlight the sensor’s potential for widespread application in robotic systems, particularly where impact resistance and cost efficiency are critical.

\subsection{Contribution}
The primary contributions of this paper are as follows:  

1. Optimization-based sensor design methodology  
   A systematic approach for modeling and optimizing six-axis force/torque sensors within constrained dimensions (40mm diameter) and force ranges is presented.  

2. Design flexibility through multiple objective functions  
   The methodology allows for design customization to achieve desired characteristics such as sensitivity, impact resistance, and reduced crosstalk.  
   
3. Incorporation of sensor placement optimization  
   The study includes the optimization of sensor placement to maximize overall performance.  

4. Implementation and experimental validation  
   The proposed sensor was implemented and integrated into a quadruped robot, with experiments conducted to verify its performance and consistency with theoretical models.  

This research introduces a robust and lightweight force/torque sensor suitable for dynamic environments such as legged robots, overcoming the limitations of existing sensors and offering broad applicability across various robotic systems.

\section{Optimal Design of Proposed Sensor}
\subsection{Principle of Proposed Sensor}

\begin{figure}[!t]\centering
	\includegraphics[width=\columnwidth]{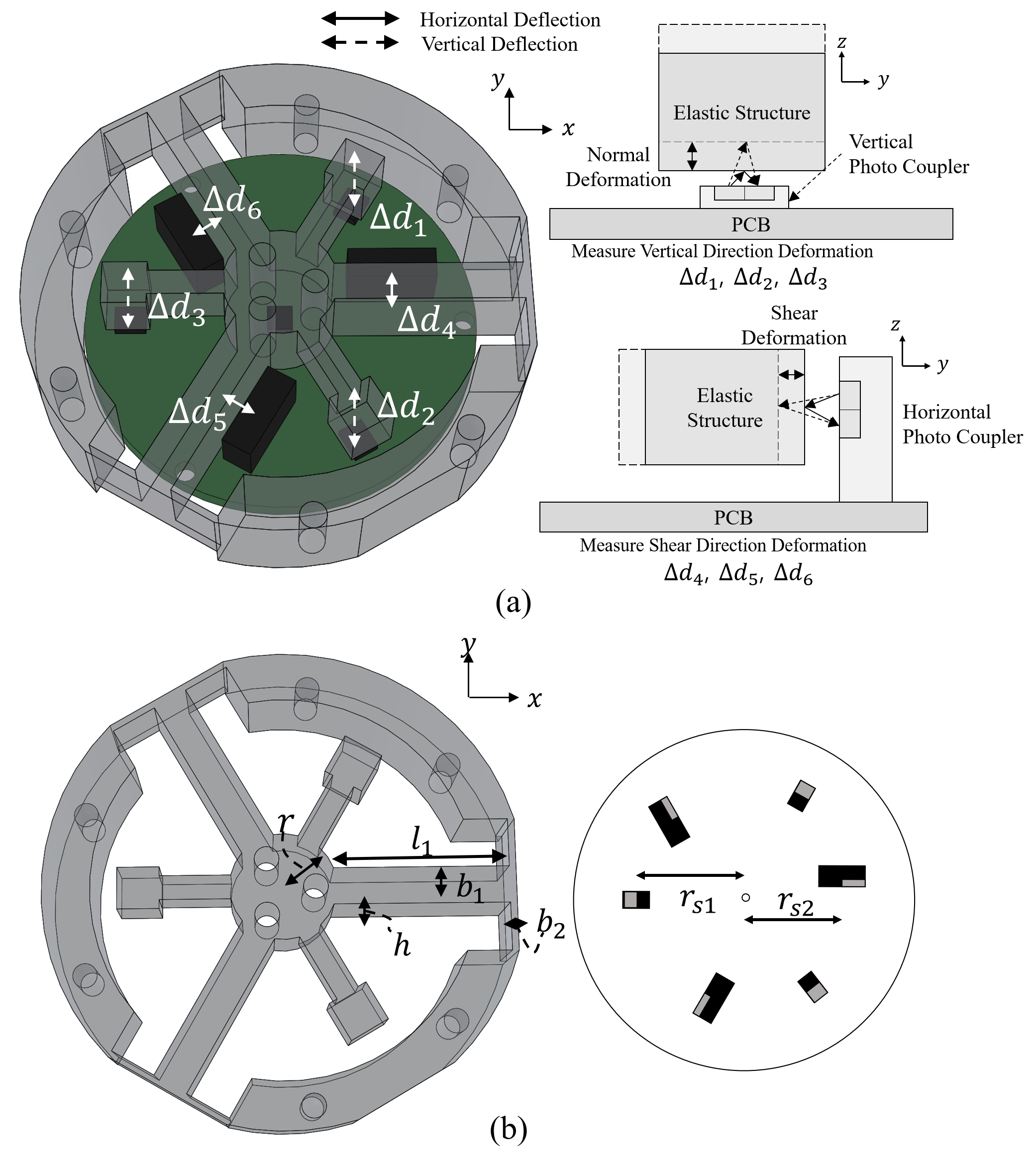}
	\caption{a) Configuration and Principle of Proposed Sensor. 3 Vertical Photo-Couplers for $F_z, M_x, M_y$ and 3 Horizontal Photo-Couplers for $F_x, F_y, M_z$ (b) Parameters of Proposed Sensor.: $r$: radius of loading table, $l_1, l_2, b_1, b_2, h$: Dimension of T-beam, and $r_{s1}$ and $r_{s2}$: Radius of Vertical Photocouplers and Horizontal Photocouplers }\label{principle}
\end{figure}

Fig.~\ref{principle} (a) illustrates the principle of the proposed sensor. $\Delta d_1$, $\Delta d_2$, and $\Delta d_3$ represent the degree of deformation due to force at the positions of the vertical photo-couplers, while $\Delta d_4$, $\Delta d_5$, and $\Delta d_6$ represent the degree of deformation due to force at the positions of the horizontal photo-couplers. The vertical photo-couplers are primarily used for measuring $F_z$, $M_x$, and $M_y$, whereas the horizontal photo-couplers are mainly utilized for measuring $F_x$, $F_y$, and $M_z$. 

The proposed sensor offers several advantages over traditional strain gauge-based methods. Unlike strain gauges, it operates using a non-contact method, and unlike capacitive sensors, it does not require an additional measurement PCB(Printed Circuit Board); a single PCB suffices. Furthermore, the sensor requires only six ADC(Analog to Digital Converter)s, reducing the number from the conventional 12 ADCs to half. Each photo-coupler measures the displacement of the spring-like component at the center, which functions as both an elastomer and a reflective surface, allowing the force to be measured. 

The elastomer in the proposed sensor employs a T-beam structure. When a force is applied to the central loading table, the elastomer deforms, enabling measurement of horizontal and vertical forces through distinct mechanisms. Horizontal forces are determined by assessing the deformation of the T-beam, while vertical forces are measured by evaluating the angle and deformation of the reflective structure associated with the loading table.

The specific photo-couplers used are the TCRT1000 and VCNT2020 from Vishay, offering measurement ranges of approximately 1mm and 0.5mm, respectively. The circuit design is highly simplified, requiring only one resistor for the diode and one resistor for the transistor to enable measurement.

Fig.~\ref{principle} (b) illustrates the structure of the sensor's elastomer, showing the dimensions of the T-beam and the positions of the sensors. The central region, referred to as the loading table, directly receives the applied force, and its radius is represented by $r$. The T-beam of the elastomer is defined by the dimensions $l_1$, $l_2$, $b_1$, $b_2$, and $h$. The positions of the vertical photo-couplers are denoted by $r_{s1}$, while those of the horizontal photo-couplers are represented by $r_{s2}$. 

In this study, these parameters were set as optimization variables. However, the position of the vertical photo-coupler was fixed at 12mm. This decision was made to address the limited space on the PCB and reduce the complexity of the optimization process, thereby improving convergence. The placement of the vertical photo-coupler at the outermost edge was chosen to enhance sensitivity, as larger displacements can be observed at the outer edges when moments around the x- and y-axes are generated by the loading table.

\subsection{Optimal Design Using Global Search Algorithm}
\begin{figure*}[!t]\centering
	\includegraphics[width=2\columnwidth]{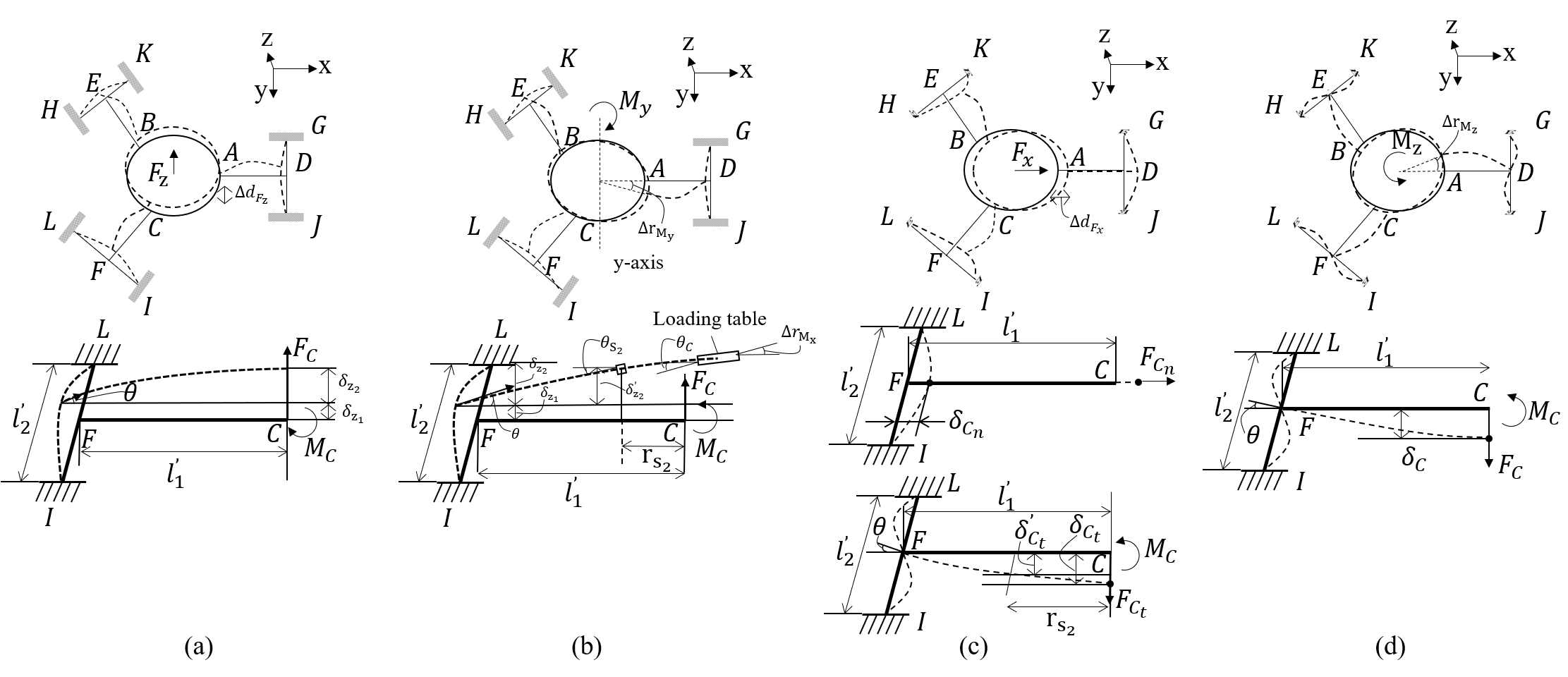}
	\caption{Elastomer deformation according to the force and moment applied along each axis:  
(a) Deformation when applying z-axis force ($F_z$)  
(b) Deformation when applying y-axis moment ($M_y$)  
(c) Deformation when applying x-axis force ($F_x$)  
(d) Deformation when applying z-axis moment ($M_z$)
    }\label{parameter}
\end{figure*}

To achieve optimal design, accurate modeling is essential. In this study, the Timoshenko beam theory was employed to develop a more precise model, as referenced in ~\cite. A model for the proposed sensor was established, and a matrix representing the relationship between force and deformation was derived. This matrix was then used to perform numerical analysis.
\begin{align}
\Delta d= \left[ \Delta d_1 \,\Delta d_2 \,\Delta d_3 \,\Delta d_4\, \Delta d_5\,  \Delta d_6 \right]^T\\
F= \left[ F_x\, F_y\, F_z\, M_x \,M_y\, M_z \right]^T\\
\Delta d = \textbf{G} F\\
\textbf{G} = \Delta d / F 
\end{align}

Eq. (1) follows the same order as shown in Fig.~\ref{principle}, where the first three terms represent the vertical photo-couplers, and the latter three terms correspond to the horizontal photo-couplers. 
Eq. (2) represents the force and moment vector, including the forces and moments along the x, y, and z axes. Eq.~(3) describes the relationship between deformation and force, where $\textbf{G}$ is defined as $\Delta d / F$, which can be interpreted as the sensitivity as Eq.~(4).

\begin{align}
\resizebox{0.9\columnwidth}{!}{$
    \textbf{G}=
    \begin{bmatrix}
    0 & 0 & -\frac{1}{k_{dFzv}} & \frac{-r_{s1}\sin(\pi/3)}{k_{rMxv}} & \frac{r_{s1}\sin(\pi/6)}{k_{rMxv}} & 0 \\ 
    0 & 0 & -\frac{1}{k_{dFzv}} & \frac{r_{s1}\sin(\pi/3)}{k_{rMxv}} & \frac{r_{s1}\sin(\pi/6)}{k_{rMxv}} & 0 \\ 
    0 & 0 & -\frac{1}{k_{dFzv}} & 0 & \frac{-r_{s1}}{k_{rMxv}} & 0 \\ 
    0 & \frac{1}{k_{dFyh}} & 0 & \frac{(h/2-c)}{k_{rMxh}} & 0 & \frac{1}{k_{dMzh}} \\ 
    \frac{-\sin(\pi/3)}{k_{dFyh}} & \frac{-\sin(\pi/6)}{k_{dFyh}} & 0 & \frac{-(h/2-c)\sin(\pi/6)}{k_{rMxh}} & \frac{-(h/2-c)\sin(\pi/3)}{k_{rMxh}} & \frac{1}{k_{dMzh}} \\ 
    \frac{\sin(\pi/3)}{k_{dFyh}} & \frac{-\sin(\pi/6)}{k_{dFyh}} & 0 & \frac{(h/2-c)\sin(\pi/6)}{k_{rMxh}} & \frac{(h/2-c)\sin(\pi/3)}{k_{rMxh}} & \frac{1}{k_{dMzh}}
    \end{bmatrix}
$}\label{3.5}
\end{align}
Here, the matrix $\textbf{G}$ of Eq.~(5) represents the relationship between forces, moments, and the deformations of each sensor. 

Each $r_{s1}$ represents the position (radius) of the vertical photo-coupler, while $h$ denotes the height of the T-beam, and $c$ indicates the difference between the midpoint height of the T-beam and the height of the horizontal photo-coupler. The parameter $k$ represents the spring constants at specific locations. 

\begin{figure}[t!]
    \centerline{\includegraphics[width=\columnwidth]{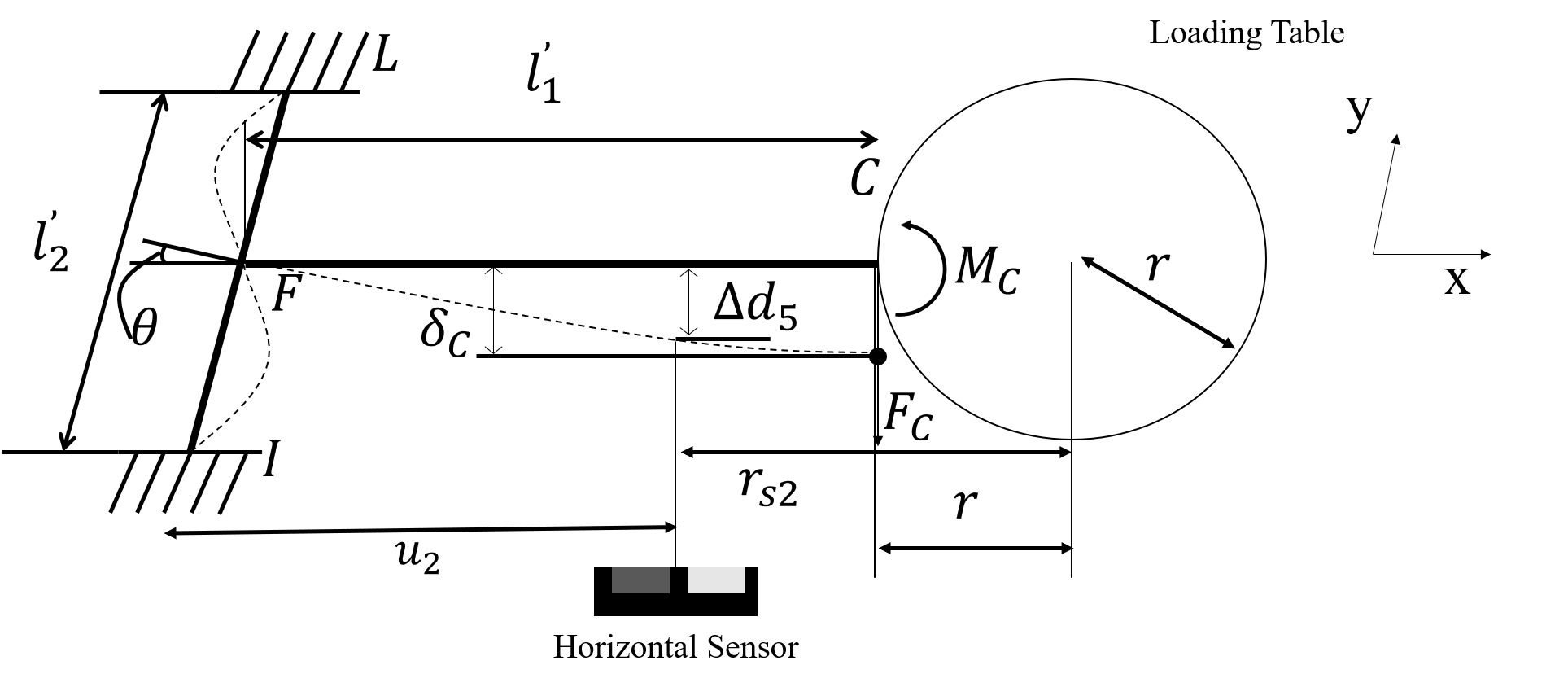}}
    \caption[Deformation of Elastomer at Horizontal Sensor]{When deformation occurs along the negative y-axis and the horizontal sensor is positioned at $r_{s2}$, the deformation measured by the horizontal sensor is \( \Delta d_5 \).

    } \label{horideform}
\end{figure}

Figure~\ref{horideform} illustrates the case where a moment $M_z$ is applied when the horizontal sensor is located at $r_{s2}$. In this example, the displacement at the original position $C$ is $\delta_C$, but due to the horizontal sensor being located at $r_{s2}$, the displacement becomes $\Delta d_5$. As such, the degree of deformation detected by the sensor varies depending on its position, necessitating a mathematical representation. Equation~\ref{3.5} incorporates the sensor's position, enabling optimization design using this equation. The notations are summarized in Table~\ref{table1}.

\begin{table}[h!]\caption{Notation of Spring Constant}
\centering \label{table1}
\begin{tabular}{cc}
\hline
$k$     & Spring Constant     \\ \hline\hline
$d$     & Displacement        \\
$r$     & Angular Deformation \\
$F$     & Force               \\
$M$     & Moment              \\
$x,y,z$ & Axis                \\
$v$     & Vertical Sensor     \\
$h$     & Horizontal Sensor   \\ \hline
\end{tabular}
\end{table}

As shown in Table~\ref{table1}, each notation corresponds to a specific parameter. For example, $k_{dFyh}$ refers to the spring constant related to the deformation caused by the y-axis force at the location of the horizontal photo-coupler, while $k_{rMxv}$ represents the spring constant associated with the angular deformation caused by the x-axis moment at the location of the vertical photo-coupler.

\subsection{Modeling}

Fig.~\ref{horideform} shows the deformation of the elastomer at the location of the horizontal sensor, which occurs when a force is applied in the -y direction. The position of the sensor can be determined using $r_{s2}$, the compensated length of the T-beam $l_1' = l_1 + b_2 / 2$, and $r$.

The force and moment applied at point C are represented as $F_C$ and $M_C$, respectively, and the resulting deformation at point C is denoted as $\delta_C$. Since the horizontal photo-coupler is not located directly at point C, the corresponding deformation at the sensor's location must be separately calculated.

Figure~\ref{parameter}.(a) illustrates how the z-axis deflection $\Delta d_{F_{z}}$ occurs due to the force along the z-axis.
\begin{equation} \Delta d_{F_{z}} =\delta _{z1} +\delta _{z2} +\theta l_{1}^{\prime }
\label{3.4}\end{equation}
As described in Equation~\ref{3.4}, $\Delta d_{F_{z}}$ is composed of the deflection due to the force $\delta_{z1}$ and the deflection due to the moment $\delta_{z2} + \theta l_{1}^{\prime}$. Here, $I_t = \beta h b_{2}^3$ represents the torsional moment of inertia, with $l_1^{\prime} = l_1 + b_2 /2$ and $l_2^{\prime} = l_2 - b_1$. The polar moment of inertia of the beam is denoted by $I$, $G$ represents the shear modulus, and $\beta$ is the torsion coefficient calculated as $(16/3 - 3.36b_{2}(1-b_{2}^4/12h^4)/h)/16$.

\begin{equation} 
\begin{cases}\theta =(F_{C} l_{1}^{\prime }-M_{C})l_{2}^{\prime }/(4GI_{t})
\\\delta _{z1} =\frac {F_{C} l_{2}^{\prime 3}}{192EI_{22}}+\frac {F_{C} l_{2}^{\prime }}{4kGS_{2}}
\\ \delta _{z2} =\frac {F_{C} {l}_{1}^{\prime 3}}{3EI_{12}}-\frac {M_{C} {l}_{2}^{\prime 2}}{2EI_{12}}+\frac {F_{C} l_{1}^{\prime }}{kGS_{1}}
\label{3.5}
\end{cases}\end{equation}
The variables $\theta$, $\delta_{z1}$, and $\delta_{z2}$ can be computed using Equation~\ref{3.5}, where $E$ represents the elastic modulus. The term $S_{2}=b_{2}h$ and $k$ is the shear coefficient, calculated as $10/(12 + b_{2}/h)$. The $I_{22} = b_{2}h^{3}/12$ is used to denote the z-axis moment of inertia. $F_C$ and $M_C$ represent the force and moment at point C, respectively.

\begin{equation} \theta _{C} =\frac {F_{C} l_{1}^{\prime 2}}{2EI_{12}}-\frac {M_{C} l_{1}^{\prime }}{EI_{12}}+\frac {(F_{C} l_{1}^{\prime }-M_{C})l_{2}^{\prime }}{4GI_{t}}=0
\label{3.6}
\end{equation}
Since the rotational angle at the beam FC is zero at C, this can be expressed geometrically as Equation ~\ref{3.6}.

\begin{equation} Fz=F_{A} +F_{B} +F_{C} =3F_{C}\label{3.7}\end{equation}
Furthermore, as outlined in Equation~\ref{3.7}, the force along the z-axis equals the sum of the forces at points A, B, and C, which can be expressed as three times $F_C$.
\begin{equation}
 \begin{aligned}
     \delta_{F_{z}}(x) = F_{C} (
    \frac{x^{2}(3l_{1}^{\prime} - x)}{6EI_{12}} 
    - \frac{a_{1} x^{2}}{2EI_{12}} 
    + \frac{(l_{1}^{\prime} - a_{1})l_{2}^{\prime}}{4GI_{t}}x \\
    + \frac{l_{2}^{\prime 3}}{192EI_{22}} 
    + \frac{4l_{1}^{\prime}S_{2} + l_{2}^{\prime}S_{1}}{4kGS_{1}S_{2}})   
 \end{aligned}
\end{equation}

Overall, this can be represented as Equation 10, where at $x=l_{1}^{\prime}$, $\Delta d_{Fz}$ can be calculated. 
Here, $k_{Fz}$ is expressed as follows.

\parbox{\linewidth}{%
   \scalebox{0.8}{
$k_{F_z} = \frac{1}{3} \Bigg( 
\frac{l_1^{\prime3}}{3 E I_{12}} 
- \frac{a_1 l_1^{\prime 2}}{2 E I_{12}} 
+ \frac{\left(l_1^{\prime}-a_1\right) l_1^{\prime} l_2^{\prime}}{4 G I_t} 
+ \frac{l_2^{\prime 3}}{192 E I_{22}} 
+ \frac{4 l_1^{\prime} S_2 + l_2^{\prime} S_1}{4 k G S_2}
\Bigg).$}}

\begin{equation} \Delta d_{F_z} =k_{F_z} F_z \label{3.9}\end{equation}

In Equation~\ref{3.9}, the spring constant relating force to deflection is presented, where $S_1=b_1 h$, $I_{12}=b_{1}h^3/12$, $F_C=F_z / 3$, and $M_C=a_1 F_C$ with $a_1=\frac{l_1^{\prime}\left(2 l_1^{\prime} G I_t+l_2^{\prime} E I_{12}\right)}{4 l_1^{\prime} G I_t+l_2^{\prime} E I_{12}}$. This is valid when $x=l_1^{\prime}$.

\begin{equation}
\begin{cases}
    \Delta r_{M_y} =\Delta r_{A_{M_y}} =2\Delta r_{C_{M_y}}\\
     \Delta r_{C_{M_y}} =\Delta d_{C_{{M_y}}}/r\\
      \Delta d_{C_{_{M_y}}} =\delta _{z1} +\delta _{z2} +\theta l_{1}^{\prime }\\
      \theta _{C} =\frac {F_{C} l_{1}^{\prime 2}}{2EI_{12}}-\frac {M_{C} l_{1}^{\prime }}{EI_{12}}+\frac {(F_{C} l_{1}^{\prime }-M_{C})l_{2}^{\prime }}{4GI_{t}}=-\Delta r_{C_{M_y}}\\
      F_{C} A_{1} =M_{C} B_{1}\\
      F_{A} A_{1} =M_{A} B_{1}
      
\end{cases}\label{3.10}\end{equation}
Figure~\ref{parameter}.(b) shows the deformation of elastomer when y-axis moment applied.
Equation~\ref{3.10} derives the deflection caused by the torque along the x-axis, where $\Delta d_{C_{_{M_y}}}$ represents the displacement at point C, which can be calculated similarly to the force along the z-axis. Additionally, the angle at C can be geometrically derived as indicated in the fourth equation of Equation~\ref{3.10}.

Here, when $ A_1=\left(\frac{l_1^{\prime 2}}{3 E I_{12}}+\frac{l_1^{\prime} l_2^{\prime}}{4 G I_t}+\frac{1}{k G S_1}\right) \lambda_1+\left(\frac{1}{4 k G S_2}+\frac{l_2^{\prime 2}}{192 E I_{22}}\right) \lambda_2+\left(\frac{l_1^{\prime 2}}{2 E I_{12}}+\frac{l_1^{\prime} l_2^{\prime}}{4 G I_t}\right) ,
 B_1=\left(\frac{l_1^{\prime}}{2 E I_{12}}+\frac{l_2^{\prime}}{4 G I_t}\right) \lambda_1+\left(\frac{l_1^{\prime}}{E I_{12}}+\frac{l_2^{\prime}}{4 G I_t}\right) , \lambda_1=l_1^{\prime} / r, \lambda_2=l_2^{\prime} / r $
equations 5 and 6 of Equation~\ref{3.10} can be derived, and using static equilibrium, this can be expressed as Equation~\ref{3.11}.

\begin{equation}
    \begin{cases} \displaystyle F_{A} =2F_{C} \\ \displaystyle My=M_{A} +F_{A} r+2(M_{C} +F_{C} r)\sin 30^{\circ } \displaystyle \end{cases}
    \label{3.11}
\end{equation} 

Consequently, the displacement at CD can be represented as follows in Equation~\ref{3.12}:
The spring constant for the torque $M_y$ can be formulated as:
\begin{multline}
    \delta _{M_y} (x)=\frac {M_y}{3(A_{1} /B_{1} +r)}\bigg(\frac {l_{2}^{\prime 3}}{192EI_{22}}+\frac {l_{2}^{\prime }}{4kGS_{2}}+\frac {x}{kGS_{1}} \\+\,\frac {x^{2}(3{l}^{\prime }_{1} -x)}{6EI_{12}}-\frac {A_1x^{2}}{2EI_{12}B_1}+\frac {(l_{1}^{\prime }-A_{1} /B_{1}){l}^{\prime }_{2}}{4GI_{t}}x\bigg) 
    \label{3.12} 
\end{multline} 
\parbox{\linewidth}{%
   \scalebox{0.70}{
$k_{M_y} = \frac{2}{3\left(A_1 / B_1 + r\right) r} \Bigg(
\frac{l_2^{l^3}}{192 E I_{22}}
+ \frac{l_2^{\prime}}{4 k G S_2}
+ \frac{2 l_1^{\prime 3} - 3 A_1 / B_1 l_1^{\prime 2}}{6 E I_{12}}
+ \frac{l_1^{\prime}}{k G S_1}
+ \frac{\left(l_1^{\prime} - A_1 / B_1\right) l_1^{\prime} l_2^{\prime}}{4 G I_t}
\Bigg)$}}


Thus, $\Delta r_{M_y} = k_{M_y} M_y$ illustrates how the displacement is directly influenced by the moment along the y-axis.

Similarly, equations for forces $F_x$ and $M_z$ can be derived as shown as Figure~\ref{parameter}.(c) and (d), enabling a comprehensive modeling of the responses to $F_z$, $M_y$, $F_x$, and $M_z$ at a distance $r$ from the center.

\begin{equation} 
\begin{cases}
    \theta =(F_{C_t} l_{1}^{\prime }-M_{C})l_{2}^{\prime }/(16EI_{21})\\
    \delta _{C_t} =\frac {F_{C_t} {l}_{1}^{\prime 3}}{3EI_{11}}-\frac {M_{C} {l}_{1}^{\prime 2}}{2EI_{11}}+\frac {F_{C_t} l_{1}^{\prime }}{kGS_{1}}+\theta l_{1}^{\prime }\\
    \delta _{C_n} =\frac {F_{C_n} {l}_{1}^{\prime 3}}{192EI_{11}}+\frac {F_{C_n} {l}^{\prime }_{1}}{ES_{1}}+\frac {F_{C_n} {l}^{\prime }_{2}}{4kGS_{2}}\\

    \end{cases}
    \label{fx1}
    \end{equation}

In the equation for $F_x$, the tangential deformation $\delta_{C_t}$ and normal deformation $\delta_{C_n}$ at $\theta$ and position $C$ can be determined using equation~\ref{fx1}. Here, $I_{11}=hb_1^3/12, I_{21}=hb_2^3/12$ was used.





    \begin{equation}
    \begin{cases}    
     \Delta d_{A} =\frac {F_{A} {l}_{2}^{\prime 3}}{192EI_{21}}+\frac {F_{A} l_{2}^{\prime }}{ES_{2}}+\frac {F_{A} l_{2}^{\prime }}{4kGS_{2}}\\
     \Delta d_{A} =\Delta d_{C} =\sqrt {\delta _{C_n}^{2}+\delta _{C_t}^{2}}\\
      \theta _{C} =\frac {F_{C_t} {l}_{1}^{\prime 2}}{2EI_{11}}-\frac {M_{C} {l}^{\prime }_{1}}{EI_{11}}+\frac {(F_{C_t} {l}^{\prime }_{1} -M_{C}){l}^{\prime }_{2}}{16EI_{21}}=0\\
      \delta _{C_n}=\tan 30^{\circ } \delta _{C_t}\\

           \end{cases}
    \end{equation}
    The deformation at point $A$ along the x-axis, $\Delta d_A$, and the deformation at point $C$, $\Delta d_C$, are identical and consist of the sum of the normal deformation and tangential deformation.
    Furthermore, since the angle $\theta_C$ at point $C$ is 0, the constraint equation can be applied. The relationship between $\delta_{C_n}$ and $\delta_{C_t}$ is governed by $\tan{30^\circ}$.

    \begin{equation}
    \begin{cases}
           F_x=2(F_{C_n} \sin 30^{\circ }+F_{C_t} \cos 30^{\circ })+F_{A}\\
       F_{A} =F_x/A_{4}, M_{C} =F_x A_{2} B_{3} /(2A_{3} A_{4} B_{2}),\\ F_{C_n} =F_x/(2A_{4}),F_{C_t} =F_x B_{3} /(2A_{3} A_{4})\\
      
           \end{cases}
           \label{3.17}
    \end{equation}
    $F_x$ is composed of $F_{C_n}$, $F_{C_t}$, and $F_A$.
    Each $F_A$, $F_x$, $M_C$, $F_{C_n}$, and $F_{C_t}$ has the same relationship as in Equation~\ref{3.17}. Additionally, each of $A_2$, $B_2$, $A_3$, $B_3$, and $A_4$ is defined as:
\[
\begin{array}{l}
A_2 = \frac{l_1^{\prime 2}}{2 I_{11}} + \frac{l_1^{\prime} l_2^{\prime}}{16 I_{21}}, \\
B_2 = \frac{l_1^{\prime}}{I_{11}} + \frac{l_2^{\prime}}{16 I_{21}}, \\
A_3 = \frac{l_1^{\prime 3}}{3 E I_{11}} + \frac{{l_1^{\prime}}^2 l_2^{\prime}}{16 E I_{21}}
+ \frac{l_1^{\prime}}{k G S_1}
- \frac{A_2 \left( \frac{l_1^{\prime 2}}{2 E I_{11}} + \frac{l_1^{\prime} l_2^{\prime}}{16 E I_{21}} \right)}{B_2}, \\
B_3 = \sqrt{3} \left( \frac{{l_1^{\prime}}^3}{192 E I_{11}} + \frac{l_1^{\prime}}{E S_1}
+ \frac{l_2^{\prime}}{4 k G S_2} \right), \\
A_4 = \left( \frac{3}{2} + \frac{\sqrt{3} B_3}{2 A_3} \right).
\end{array}
\]
This structure defines the relationships and dependencies among the variables.

  By solving the above equations, $\delta_{F_x}$ can be determined as a function of $x$, i.e., based on position.

    \begin{equation}
    \resizebox{0.8\columnwidth}{!}{
    $\begin{cases}
 \delta _{F_x} (x)=\sqrt {\delta _{C_n} (x)^{2}+\delta _{C_t} (x)^{2}} =\delta _{C_t} (x)/cos30^{\circ }\\
       \delta _{C_n} (x)=\frac {F_{C_n} l_{2}^{3}}{192EI_{21}}+\frac {F_{C_n} x}{ES_{1}}+\frac {F_{C_n} x}{4kGS_{2}}\\
       \delta _{C_t} (x)=\frac {F_{C_t} x^{2}(3l_{1}^{\prime }-x)}{6EI_{11}}-\frac {M_{C} x^{2}}{2EI_{11}}+\frac {(F_{C_t} l_{1}^{\prime }-M_{C})l_{2}^{\prime }}{16EI_{21}}x+\frac {F_{C_t} x}{4kGS_{1}}\\
    \end{cases}$}
    \label{3.18}
    \end{equation}

    The deformation equation expressed as $x$ in Equation~\ref{3.18} can be used to calculate $\delta_{F_x}(x)$, which in turn is useful for determining $\Delta d_{F_x}$. Here, $k_{F_x}$ represents the condition when $x$ equals $l_1^{\prime}$.
    \begin{equation}
       \Delta d_{F_x} =k_{F_x} F_x
    \end{equation}
where,
\parbox{\linewidth}{%
   \scalebox{0.8}{
$k_{F_x} = \frac{B_3}{2 A_3 A_4 \cos 30^{\circ}}
\left(\frac{l_1^{\prime 3}}{3 E I_{11}}
- \frac{A_2 l_1^{\prime 2}}{2 E I_{11} B_2} \right.
 + \frac{{l_1^{\prime}}^2 l_2^{\prime} - A_2 l_1^{\prime} l_2^{\prime} / B_2}{16 E I_{21}}
+ \frac{l_1}{4 k G S_1} \bigg)$
}}.
 



To calculate the deformation of $M_z$, the relationships between $F_C$, $M_C$, $\theta_C$, and $\delta_C$ are utilized.

\begin{equation}\begin{cases}
    3F_{C} r+3M_{C} =M_z\\
    \theta _{C} =\frac {F_{C} l_{1}^{2}}{2EI_{11}}-\frac {M_{C} l_{1}}{EI_{11}}+\frac {(F_{C} l_{1}^{\prime }-M_{C})l_{2}^{\prime }}{16EI_{21}}=-\delta _{C} /r\\
    \delta _{C} =\frac {F_{C} {l}_{1}^{\prime 3}}{3EI_{11}}-\frac {M_{C} {l}_{1}^{\prime 2}}{2EI_{11}}+\frac {(F_{C} l_{1}^{\prime }-M_{C}){l}^{\prime }_{1} l_{2}^{\prime }}{16EI_{21}}+\frac {F_{C} {l}^{\prime }_{1}}{kGS_{1}}\\
         \end{cases}\label{3.20}
    \end{equation}
The value of $\theta_C$ can be obtained from $\delta_C$ by dividing it by the radius $r$ of the loading table, as shown in Equation~\ref{3.20}.

    \begin{equation}
    \begin{cases}
    F_{C} =M_z/\left({3\left({r+\frac {1}{a_{2}}}\right)}\right)\\
    M_{C} =M_z/(3(a_{2} r+1))\\
         \end{cases}
    \end{equation}
    The relationship between $F_C$, $M_C$, and $M_z$ is expressed through the parameter $a_2$, which is defined as follows:

   \parbox{\linewidth}{%
   \scalebox{0.8}{
    $a_2=\frac{l_1^{\prime} /\left(E I_{11}\right)+l_2^{\prime} /\left(16 E I_{21}\right)+\left(l_1^{\prime} /\left(2 E I_{11}\right)+l_2^{\prime} /\left(16 E I_{21}\right)\right) \lambda_1}{l_1^{\prime 2} /\left(2 E I_{11}\right)+l_1^{\prime} l_2^{\prime} /\left(16 E I_{21}\right)+\left(l_1^{\prime 2} /\left(3 E I_{11}\right)+l_1^{\prime} l_2^{\prime} /\left(16 E I_{21}\right)+1 /\left(k G S_1\right)\right) \lambda_1}$}}
    \begin{equation}
        \resizebox{0.85\columnwidth}{!}{
    $\begin{cases}
    \delta _{M_z} (x)=F_{C} \bigg(\frac {x^{2}}{6EI_{11}}(-x+3{l}^{\prime }_{1})-\frac {a_{2} x^{2}}{2EI_{11}}\\+\frac {(l_{1}^{\prime }-a_{2})l_{2}^{\prime }}{16EI_{21}}x+\frac {x}{kGS_{1}}\bigg)\\
    \Delta r_{M_z} =k_{M_z} M_z\\
    k_{M_z}=\frac{1}{3r \left(r+1/ a_2\right)}\left(\frac{l_1^{\prime 3}}{3 E I_{11}}-\frac{l_1^{\prime 2} a_2}{2 E I_{11}}+\frac{l_1^2 l_2^{\prime}-l_1^{\prime} l_2^{\prime} a_2}{16 E I_{21}}+\frac{l_1^{\prime}}{k G S_1}\right)
\end{cases}$ }\label{3.22}
    \end{equation}




Equation~\ref{3.22} can be expressed as a function of $x$, allowing the calculation of $\delta_{M_z}(x)$ and determining $k_{M_z}$ when $x$ equals $l_1^{\prime}$. However, to achieve optimization, the sensor positions also need to be optimized. Therefore, it is necessary to calculate the spring constant at each sensor position.

Using this approach, the spring constant corresponding to each sensor position can be determined as shown in Equation~\ref{3.23}. These values are then utilized to compute the deformation and the matrix relating forces/torques. Equation~\ref{3.5} represents the matrix that relates deformation to applied forces and torques based on the spring constants at the optimized sensor positions.
    

\begin{equation}  \resizebox{1\columnwidth}{!}{
$\begin{array}{l}
    u = {l_1}^{'} + r - r_{s1} \\
k_{dFzv} = 1/(\frac{1}{3} \left( \frac{u^2 (3{l_1}^{\prime} - u)}{6EI_{12}} - \frac{a_{1} u^2}{2EI_{12}} + \frac{({l_1}^{\prime} - a_{1}) {l_2}^{\prime} u}{4GI_t} + \frac{{{l_2}^{\prime}}^3}{192EI_{22}} + \frac{4{l_1}^{\prime} S_{2} + {l_2}^{\prime} S_{1}}{4k G S_{1} S_{2}} \right)) \\
k_{rMyv} = r/(\frac{2}{3\left(\frac{A_{1}}{B_{1}} + r\right)} \left( \frac{{{l_2}^{\prime}}^3}{192EI_{22}} + \frac{{l_2}^{\prime}}{4k G S_{2}} + \frac{u}{k G S_{1}} + \frac{u^2 (3{l_1}^{\prime} - u)}{6EI_{12}} - \frac{A_{1}}{B_{1}} \frac{u^2}{2EI_{12}} + \frac{({l_1}^{\prime} - \frac{A_{1}}{B_{1}}) {l_2}^{\prime} u}{4GI_t} \right))\\

u = {l_1}^{'} + r - r_{s2} \\
k_{dFxh} = 1/({\frac{B3}{2A3 A4 \cos(\pi/6)} \left( \frac{u^2 (3{l_1}^{\prime} - u)}{6EI_{11}} - \frac{A_{2} u^2}{2EI_{11} B2} + \frac{{l_1}^{\prime} {l_2}^{\prime} u - A_{2} u {l_2}^{\prime} / B2}{16EI_{21}} + \frac{u}{4k G S_{1}} \right)}) \\
k_{dMzh} = 1/({\frac{1}{3(r + \frac{1}{a_{2}})} \frac{u^2 (3{l_1}^{\prime} - u)}{6EI_{11}} - \frac{1}{3(a_{2} r + 1)} \frac{u^2}{2EI_{11}} + \left( \frac{1}{3(r + \frac{1}{a_{2}})} {l_1}^{\prime} - \frac{1}{3(a_{2} r + 1)} \right) \frac{u {l_2}^{\prime}}{16EI_{21}} + \frac{1}{3(r + \frac{1}{a_{2}})} \frac{{l_1}^{\prime}}{k G S_{1}}}) \\
k_{rMyh} = r/({\frac{2}{3\left(\frac{A_{1}}{B_{1}} + r\right)} \left( \frac{{{l_2}^{\prime}}^3}{192EI_{22}} + \frac{{l_2}^{\prime}}{4k G S_{2}} + \frac{u}{k G S_{1}} + \frac{u^2 (3{l_1}^{\prime} - u)}{6EI_{12}} - \frac{A_{1}}{B_{1}} \frac{u^2}{2EI_{12}} + \frac{({l_1}^{\prime} - \frac{A_{1}}{B_{1}}) {l_2}^{\prime} u}{4GI_t} \right)})
\end{array}$}
\label{3.23}
\end{equation}

The term $\Delta d$ denotes the deformations, $\overline{\textbf{G}}$ refers to the regulated matrix, and $\textbf{R}_s^{-1}$ is the regulation matrix, which is defined as:
\begin{equation}
\begin{aligned}
\overline{{\textbf{G}}} = \textbf{R}_s^{-1} \textbf{G}
\end{aligned}
 \label{3.24}
\end{equation}

\begin{equation}
\resizebox{0.8\columnwidth}{!}{$
\begin{aligned}
\min\limits_{\textbf{G}\in\mathbb{R}^{6\times6}}  \text{Function}(\overline{\textbf{G}} ) \quad
\text {s.t} 
\begin{cases} 
\displaystyle{1 \textrm {mm}\le l_{1} \le 20 \textrm {mm}}, \\
\displaystyle {11 \textrm {mm}\le l_{2} \le 30 \textrm {mm}}, \\
\displaystyle {1 \textrm {mm}\le b_{1} \le 10 \textrm {mm}}, \\
\displaystyle {0.5 \textrm {mm}\le b_{2} \le 1 \textrm {mm}}, \\
\displaystyle {1 \textrm {mm}\le h \le 15 \textrm {mm}}, \\
\displaystyle {1 \textrm {mm}\le r\le 8 \textrm {mm}}, \\
\displaystyle {2 \textrm {mm}\le r_{s2}\le 15 \textrm {mm}}, \\
 -l_{1} +3b_{1} < 0, \\ 
 -l_{2} +3b_{2} < 0, \\
-0.02+\sqrt{{(r+l_{1}+b_{2})}^2+(l_{2}/2)^2}<0, \\
\sigma_{bend}-\sigma_{allowable}<0, \\
\sigma_{torsion}-\sigma_{allowable}<0.
\end{cases}
\end{aligned}$}
\label{3.25}
\end{equation}



In equation~\ref{3.25}, the constraints for \( l_1 \), \( l_2 \), \( b_1 \), \( b_2 \), \( h \), \( r \), and \( r_{s2} \) were set with broad limits. Given that the diameter is 40 mm, the constraints were defined to ensure compliance, including \( -0.02+\sqrt{{(r+l_{1}+b_{2})}^2+(l_{2}/2)^2}<0 \). Additionally, considering the allowable stress, the modeling process incorporated the accuracy conditions from equations~\ref{3.6} to \ref{3.22}, specifically the constraints \( -l_{1} +3b_{1} < 0 \) and \( -l_{2} +3b_{2} < 0 \).


\begin{align}
\resizebox{0.8\columnwidth}{!}{
$
\mathbf{R}_s^{-1} = \text{diag} \left[ 
    \frac{100\%}{520 \, \textrm{N}}, 
    \frac{100\%}{520 \, \textrm{N}}, 
    \frac{100\%}{520 \, \textrm{N}}, 
    \frac{100\%}{15.6 \, \textrm{N} \cdot \textrm{m}}, 
    \frac{100\%}{15.6 \, \textrm{N} \cdot \textrm{m}}, 
    \frac{100\%}{15.6 \, \textrm{N} \cdot \textrm{m}} 
\right]
$
}\label{3.26}
\end{align}

Equation~\ref{3.26} is a normalization matrix, where the maximum target force of \( 520 \)N and moment of 15.6N\( \cdot  \)m were set as 100\%, and the normalization matrix was applied accordingly in equation~\ref{3.24}.

\begin{equation}
\begin{split}
   & \hspace{0cm} f=\text{Cond}(\overline{\textbf{G}}) \\
   & \hspace{0cm} f=1/\Vert \overline{\textbf{G}} \Vert_2\\
   & \hspace{0cm} f=1/\Vert \overline{\textbf{G}} \Vert_F\\
   & \hspace{0cm} f=1/\Vert \overline{\textbf{G}} \Vert_*\\
   & \hspace{0cm} f=\text{Cond}(\overline{\textbf{G}})/\Vert \overline{\textbf{G}}\Vert_F\\
   & \hspace{0cm} f=\text{Cond}(\overline{\textbf{G}})\Vert \overline{\textbf{G}} \Vert_2\\
   & \hspace{0cm} f=\text{Cond}(\overline{\textbf{G}})\Vert \overline{\textbf{G}} \Vert_2/\Vert \overline{\textbf{G}}\Vert_F\\
   & \hspace{0cm} f=\text{Cond}(\overline{\textbf{G}})\Vert \overline{\textbf{G}} \Vert_2/\Vert \overline{\textbf{G}}\Vert_*\\
   & \hspace{0cm} f=\text{Cond}(\overline{\textbf{G}})\Vert \overline{\textbf{G}} \Vert_2/{\Vert \overline{\textbf{G}}\Vert_*}^2
\end{split}
\label{3.27}
\end{equation}

 \begin{figure}[h]
    \centerline{\includegraphics[width=\columnwidth]{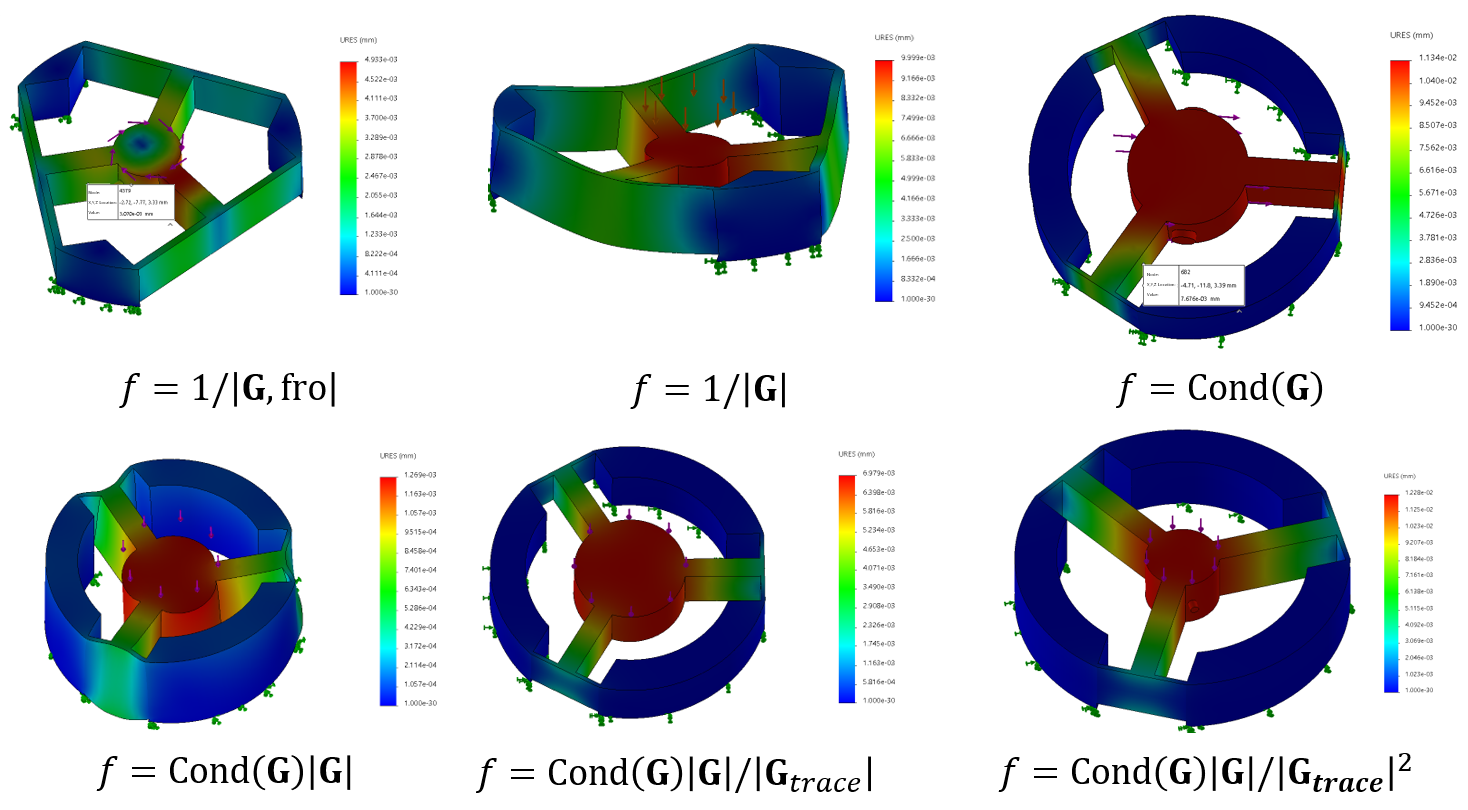}}
    \caption[Sensor Configuration by Objective Function using FEM]{Sensor Configuration by Objective Function using FEM
    } \label{femfunc}
\end{figure}
\begin{table*}[!h]
\centering
\caption{Optimized Variables, Condition Numbers, and Deformation Analysis of Various Functions}
\resizebox{\textwidth}{!}{
\begin{tabular}{ccccccccccccc}
\hline
$function$    & $l_1$(mm) & $l_2$(mm) & $b_1$(mm) & $b_2$(mm) & $h$(mm) & $r$(mm) & $r_{s2}$(mm) & $\text{Cond}(\overline{\textbf{G}})$ & $F_z$ 100N(mm) & $F_x$ 100N(mm) & $M_z$ 1N$\cdot$m(mm) & $M_y$ 1N$\cdot$m(rad) \\ \hline \hline
$\text{Cond}(\overline{\textbf{G}})$                & 10.97   & 11.00    & 3.656    & 0.5009   & 7.536  & 7.761  & 12.64   & 47.4971    & 5.583e-03 & 7.676e-03 & 2.052e-03 & 1.103e-03 \\ 
$\frac{1}{\Vert \overline{\textbf{G}} \Vert_2}$     & 9.139   & 27.45    & 3.046    & 0.9714   & 6.658  & 4.437  & 15.00   & 1558.73    & 9.999e-03 & 8.373e-03 & 1.247e-03 & 9.755e-04 \\ 
$\frac{1}{\Vert \overline{\textbf{G}} \Vert_F}$     & 9.139   & 27.45    & 3.046    & 0.9714   & 6.658  & 4.437  & 8.237   & 269.821    & 1.000e-03 & 1.673e-02 & 3.070e-03 & 9.755e-04 \\ 
$\frac{1}{\Vert \overline{\textbf{G}} \Vert_*}$     & 14.24   & 11.00    & 3.039    & 0.5535   & 6.690  & 4.437  & 8.739   & 47.8550    & 1.195e-02 & 1.655e-02 & 5.460e-03 & 7.325e-04 \\ 
$\frac{\text{Cond}(\overline{\textbf{G}})}{\Vert \overline{\textbf{G}} \Vert_2}$ & 14.24   & 11.00    & 3.039    & 0.5535   & 6.690  & 4.437  & 10.18   & 47.8485    & 1.195e-02 & 1.572e-02 & 5.646e-03 & 7.309e-04 \\ 
$\frac{\text{Cond}(\overline{\textbf{G}})}{\Vert \overline{\textbf{G}} \Vert_F}$ & 14.24   & 11.00    & 3.039    & 0.5535   & 6.690  & 4.437  & 8.913   & 47.8561    & 1.195e-02 & 1.655e-02 & 5.846e-03 & 7.309e-04 \\ 
$\frac{\text{Cond}(\overline{\textbf{G}})}{\Vert \overline{\textbf{G}} \Vert_*}$ & 14.24   & 11.00    & 3.039    & 0.5535   & 6.690  & 4.437  & 8.716   & 47.8548    & 1.195e-02 & 1.653e-02 & 5.867e-03 & 7.325e-04 \\ 
$\text{Cond}(\overline{\textbf{G}})\Vert \overline{\textbf{G}} \Vert_2$ & 10.92   & 11.00    & 3.640    & 0.7049   & 15.00  & 7.603  & 14.80   & 50.8431    & 1.269e-03 & 2.232e-03 & 8.402e-04 & 8.571e-05 \\ 
$\frac{\text{Cond}(\overline{\textbf{G}})\Vert \overline{\textbf{G}} \Vert_2}{\Vert \overline{\textbf{G}} \Vert_F}$ & 10.69   & 11.00    & 2.934    & 0.5379   & 7.176  & 8.000  & 11.80   & 47.9025    & 6.979e-03 & 1.207e-02 & 3.864e-03 & 3.816e-04 \\ 
$\frac{\text{Cond}(\overline{\textbf{G}})\Vert \overline{\textbf{G}} \Vert_2}{\Vert \overline{\textbf{G}} \Vert_*}$ & 10.99   & 11.00    & 2.938    & 0.5385   & 7.157  & 7.701  & 11.50   & 47.9085    & 6.979e-03 & 1.207e-02 & 3.864e-03 & 3.964e-04 \\
$\frac{\text{Cond}(\overline{\textbf{G}})\Vert \overline{\textbf{G}} \Vert_2}{{\Vert \overline{\textbf{G}} \Vert_F}^2}$ & 14.24   & 11.00    & 3.039    & 0.5535   & 6.690  & 4.437  & 8.908   & 47.8561    & 1.228e-02 & 1.735e-02 & 6.422e-03 & 7.607e-04 \\
$\frac{\text{Cond}(\overline{\textbf{G}})\Vert \overline{\textbf{G}} \Vert_2}{{\Vert \overline{\textbf{G}} \Vert_*}^2}$ & 14.24   & 11.00    & 3.039    & 0.5535   & 6.690  & 4.437  & 8.716   & 47.8548    & 1.228e-02 & 2.125e-02 & 6.449e-03 & 7.607e-04 \\ 
$\frac{\text{Cond}(\overline{\textbf{G}})\Vert \overline{\textbf{G}} \Vert_2}{\Vert \overline{\textbf{G}} \Vert_F\Vert \overline{\textbf{G}} \Vert_*}$ & 14.24   & 11.00    & 3.039    & 0.5535   & 6.690  & 4.437  & 8.800   & 47.8554    & 1.228e-02 & 2.125e-02 & 6.449e-03 & 7.607e-04 \\ \hline
\end{tabular}
}
\label{combined_table}
\end{table*}

The meaning of each function is as follows:

$\text{Cond}(\overline{\textbf{G}})$ represents the ratio of the maximum to minimum singular values, and when it is close to 1, it indicates better isotropy and reduced sensor error. This ensures that no particular mode dominates, leading to more balanced performance across all modes.

$\Vert\overline{\textbf{G}}\Vert_2$ represents the maximum singular value of the sensor matrix, and minimizing the objective function increases $\Vert\overline{\textbf{G}}\Vert_2$, thereby amplifying the maximum singular value. This implies that certain modes are emphasized.

$\Vert \overline{\textbf{G}} \Vert_F$, also known as the Frobenius norm or the 2-norm, is the square root of the sum of all the squared elements of the matrix. This norm reflects the overall magnitude of the matrix, which differs slightly from singular values but is useful for representing the general scale of the matrix. Minimizing the objective function increases $\Vert \overline{\textbf{G}}\Vert_F$, which in turn maximizes the overall sensitivity of the sensor.

$\Vert \overline{\textbf{G}}\Vert_*$ represents the sum of singular values, maximizing the output across all modes, and is another type of norm indicating matrix size. Minimizing the objective function finds a combination that increases the overall sum of the singular values, although the matrix size grows.

Thus, the above four combinations were considered. Minimizing $\text{Cond}(\overline{\textbf{G}})$/$\Vert\overline{\textbf{G}}\Vert_2$ leads to the minimization of $\text{Cond}(\overline{\textbf{G}})$ while maximizing $\Vert\overline{\textbf{G}}\Vert_2$. This also minimizes 1/$\Vert\overline{\textbf{G}}^{-1}\Vert_2$, thereby maximizing $\Vert\overline{\textbf{G}}^{-1}\Vert_2$. Similarly, minimizing $\text{Cond}(\overline{\textbf{G}})$/$\Vert\overline{\textbf{G}}\Vert_F$ minimizes $\text{Cond}(\overline{\textbf{G}})$ while maximizing the overall size of the matrix. Likewise, minimizing $\text{Cond}(\overline{\textbf{G}})$/$\Vert\overline{\textbf{G}}\Vert_*$ minimizes $\text{Cond}(\overline{\textbf{G}})$ while maximizing the sum of the singular values, although this aims to maximize the sum rather than distribute outputs evenly across all modes.

From the sensor's perspective, a more effective objective function would minimize $\text{Cond}(\overline{\textbf{G}})$ and $\Vert\overline{\textbf{G}}\Vert_2$, while maximizing sensor sensitivity. Hence, $\text{Cond}(\overline{\textbf{G}})$$\Vert\overline{\textbf{G}}\Vert_2$ was placed in the numerator, and the sensitivity-representing Frobenius norm and trace norm were used in the denominator for comparison.

 \begin{figure*}[!h]
 \centering
 \includegraphics[width=2\columnwidth]{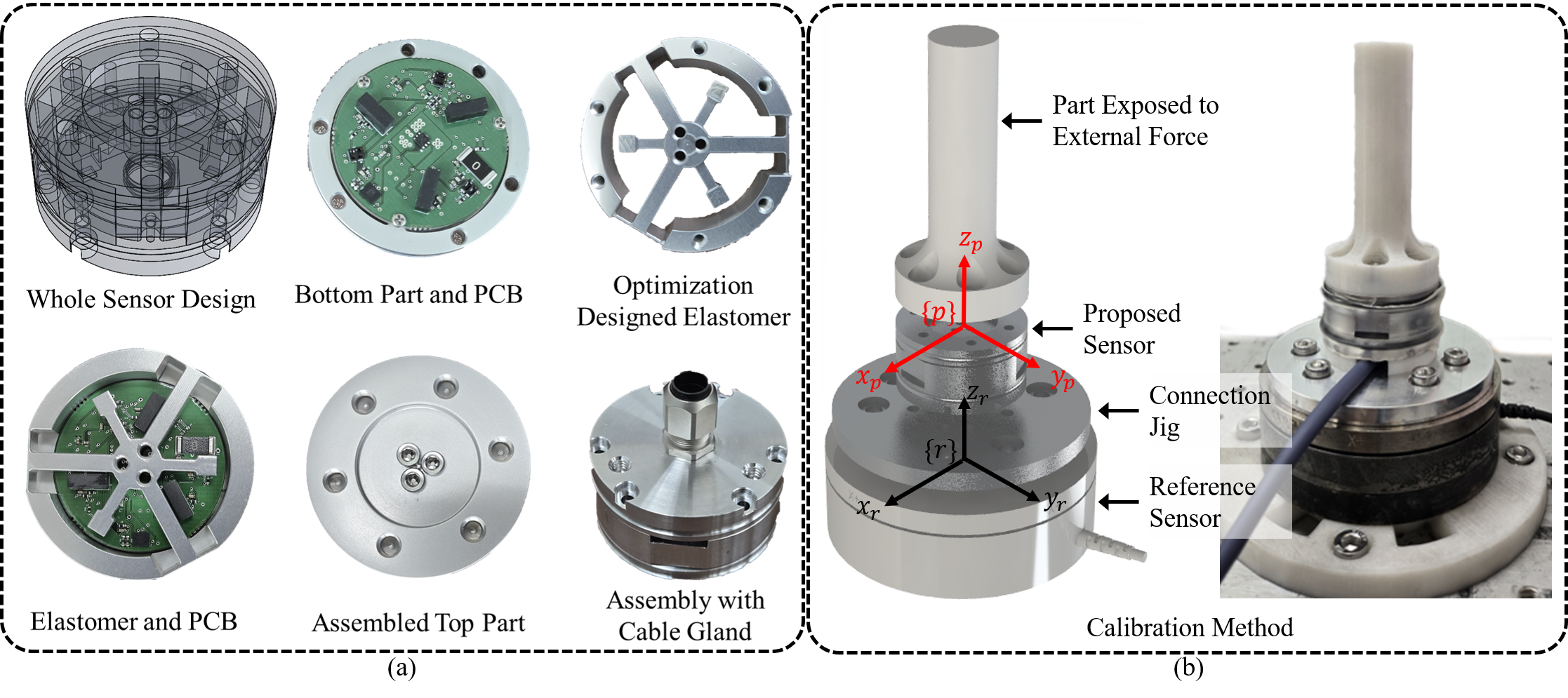}
    \caption[Sensor Configuration by Objective Function using FEM]{Sensor Configuration by Objective Function using FEM: (a) Each Part of Sensors and Assembly (b) Calibration Method(Reference Sensor: ATI MINI85)
    } \label{sensorconfig}
\end{figure*}

Since $\text{Cond}(\overline{\textbf{G}})$ is defined as $\Vert\overline{\textbf{G}}\Vert_2 \Vert\overline{\textbf{G}}^{-1}\Vert_2$, minimizing $\text{Cond}(\overline{\textbf{G}}) \Vert\overline{\textbf{G}}\Vert_2$ represents $\Vert\overline{\textbf{G}}\Vert_2^2 \Vert\overline{\textbf{G}}^{-1}\Vert_2$. Therefore, the effectiveness of a matrix size norm is limited when its exponent is 1, and it becomes more impactful when the exponent is 2. At this point, the Frobenius norm and trace norm can be compared. The Frobenius norm reflects the overall magnitude, while the trace norm sums the singular values, making them fundamentally different. An increase in the trace norm does not necessarily mean an even increase in output across all modes, but because the maximum singular value norm and $\text{Cond}(\overline{\textbf{G}})$ are present in the numerator, an increase in the trace norm can lead to more balanced outputs. Similarly, increasing the Frobenius norm enhances overall sensitivity but it can not make balanced outputs, and since my goal is to improve sensitivity while minimizing errors and avoiding excessive force in specific directions, I selected the function $\text{Cond}(\overline{\textbf{G}})\Vert\overline{\textbf{G}}\Vert_2$/${\Vert\overline{\textbf{G}}\Vert_*}^2$

Each of the objective functions was compared using $\text{Cond}(\overline{\textbf{G}})$ and finite element method (FEM) results. The findings are summarized in the following table~\ref{condnum11} and table~\ref{condnum111}.

Analyzing Function:

The condition number $\text{Cond}(\overline{\textbf{G}})$ alone does not take into account the magnitude of the singular values. This means that it only serves to ensure that the difference between the singular values is minimized, without contributing to maximizing the overall sensitivity. The term $1/\Vert \overline{\textbf{G}} \Vert_2$ primarily affects the maximum singular value, but does not influence the minimum singular value, thus acting in only one mode or direction rather than improving overall characteristics. On the other hand, $1/\Vert \overline{\textbf{G}} \Vert_F$ reflects the overall size and affects the sensitivity, but does not address the difference between singular values, merely capturing the total size without adequately influencing their magnitudes.

Furthermore, $1/\Vert \overline{\textbf{G}} \Vert_*$ maximizes the trace norm, which corresponds to the sum of the singular values. This could lead to an imbalance where one singular value becomes disproportionately large, thereby enhancing certain directions.

By using $\text{Cond}(\overline{\textbf{G}})\Vert \overline{\textbf{G}} \Vert_2$, the difference between the maximum and minimum singular values is reduced while also decreasing the maximum singular value. This helps to prevent strong influence in a specific direction. However, this may negatively impact sensitivity. Since my goal is to also maximize sensitivity, compensation using the Frobenius norm or the trace norm is necessary.

Comparison with $\text{Cond}(\overline{\textbf{G}})\Vert \overline{\textbf{G}} \Vert_2 / \Vert \overline{\textbf{G}} \Vert_F$: The term $\text{Cond}(\overline{\textbf{G}})\Vert \overline{\textbf{G}} \Vert_2$ minimizes $\text{Cond}(\overline{\textbf{G}})$ to reduce errors while increasing the overall matrix size, thereby considering sensitivity in specific directions. Placing the squared Frobenius norm in the denominator increases the overall size of the sensor but does not yield optimal results in terms of singular values. However, Frobenius norm can not balance singular values. 

Conclusion: I selected $\text{Cond}(\overline{\textbf{G}})\Vert \overline{\textbf{G}} \Vert_2 / {\Vert \overline{\textbf{G}} \Vert_*}^2$. Since the trace norm represents the sum of singular values, it may be more suitable for maximizing sensitivity in specific directions. However, even if the trace norm increases, singular values may not be uniformly distributed. Despite this, with both $\text{Cond}(\overline{\textbf{G}})$ and $\Vert \overline{\textbf{G}} \Vert_2$ present, a more uniform distribution of singular values is possible.

\section{Manufacturing and Calibration}
\begin{figure*}[!h]
    \centering
    \includegraphics[width=2\columnwidth]{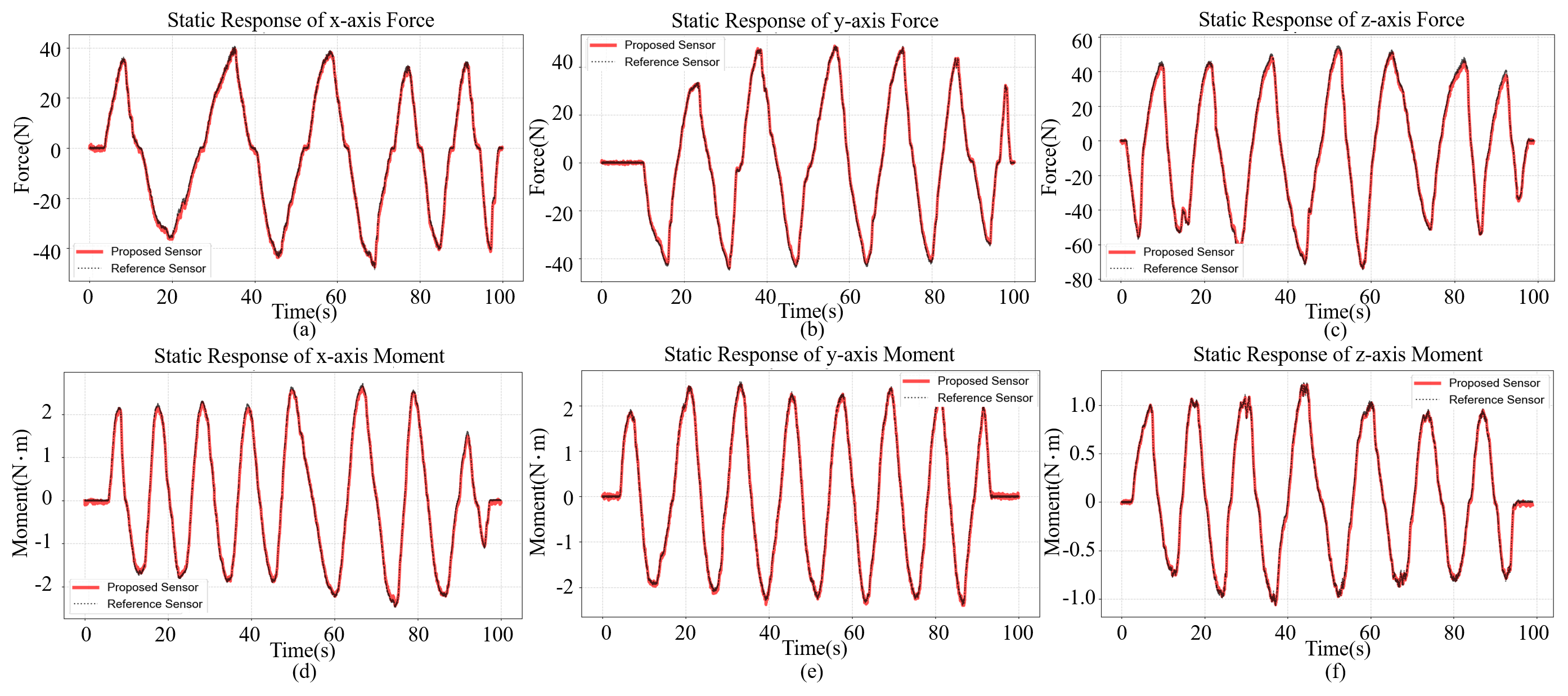} 
    \caption[Calibration Result]{Static Test of Calibration Result (a) Result of x-axis Force (b) Result of y-axis Force (c) Result of z-axis Force (d) Result of x-axis Moment (e) Result of y-axis Moment (f) Result of z-axis Moment}
    \label{calres}
\end{figure*}

The sensor was fabricated through machining and silver anodizing, utilizing the optimized values derived earlier. The material used is AL7075-T6, an aluminum alloy known for its low silicon content, which minimizes hysteresis, and its high strength properties. 

The measurement range of the sensor is summarized in Table~\ref{sensrange}, with a maximum measurement capacity of approximately 1900 N.

\begin{table}[!h]\caption[Sensor Range]{Sensor Measuring Range}\begin{center}
\begin{tabular}{cc}
\hline
Force/Moment       & Sensing Range             \\ \hline\hline
$F_x$   & -620N$\sim$+620N     \\ 
$F_y$   & -590N$\sim$+590N       \\ 
$F_z$   & -1965N$\sim$+1965N     \\ 
$M_x$ & -13.7N$\cdot$m$\sim$+13.7N$\cdot$m \\ 
$M_y$ & -13.6N$\cdot$m$\sim$+13.6N$\cdot$m \\ 
$M_z$ & -19.6N$\cdot$m$\sim$+19.6N$\cdot$m \\ \hline
\end{tabular}\end{center}\label{sensrange}
\end{table}
The sensor consists of a total of four parts. One is the bottom part, which secures the sensor in place. Another is the printed circuit board (PCB), which was designed based on the optimized sensor positions. The PCB includes a CAN transceiver and is equipped with an STM H7 series MCU, enabling 16-bit ADC measurements to be performed internally and transmitted via CAN communication at speeds of up to 5 Mbps.

Additionally, the sensor includes an elastomer designed with optimized geometry, serving both as a spring and a reflective surface. The final component is the top part, which applies force to the sensor. This configuration comprises three mechanical components and one electronic component.


The calibration process utilized the ATI-IA MINI-85 as the reference sensor, which has a resolution of 0.08N–0.32N for force and 0.0033N$\cdot$m–0.013N$\cdot$m for torque, with a sampling rate exceeding 5 kHz. The MINI-85 exhibits an approximate full-scale error of 2\%. Given its maximum sensing range of 3800N, it is suitable for measuring the proposed sensor~\cite{kim2024compact}.


Calibration was performed using TCRT1000 and VCNT2020 sensors by directly measuring the distance and analyzing the corresponding voltage changes. A 7th-order polynomial curve fitting was applied to the experimental data using the Least Squares Method. The reason for using a 7th-order polynomial is to address the nonlinearity of the photocoupler sensor, which becomes significant as the distance changes, thereby improving the accuracy of the nonlinear fitting.

Figure~\ref{calres} shows the results of the calibration after applying static forces and torques. Each experiment was conducted over a duration of 100 seconds. The percentage error, RMSE, nonlinearity, and hysteresis obtained from the experiments are summarized in Table~\ref{combined_error_analysis}. The maximum error was found to be as low as 0.88\%, and the hysteresis remained below 2\%.



\begin{figure}[!t]
    \centerline{\includegraphics[width=\columnwidth]{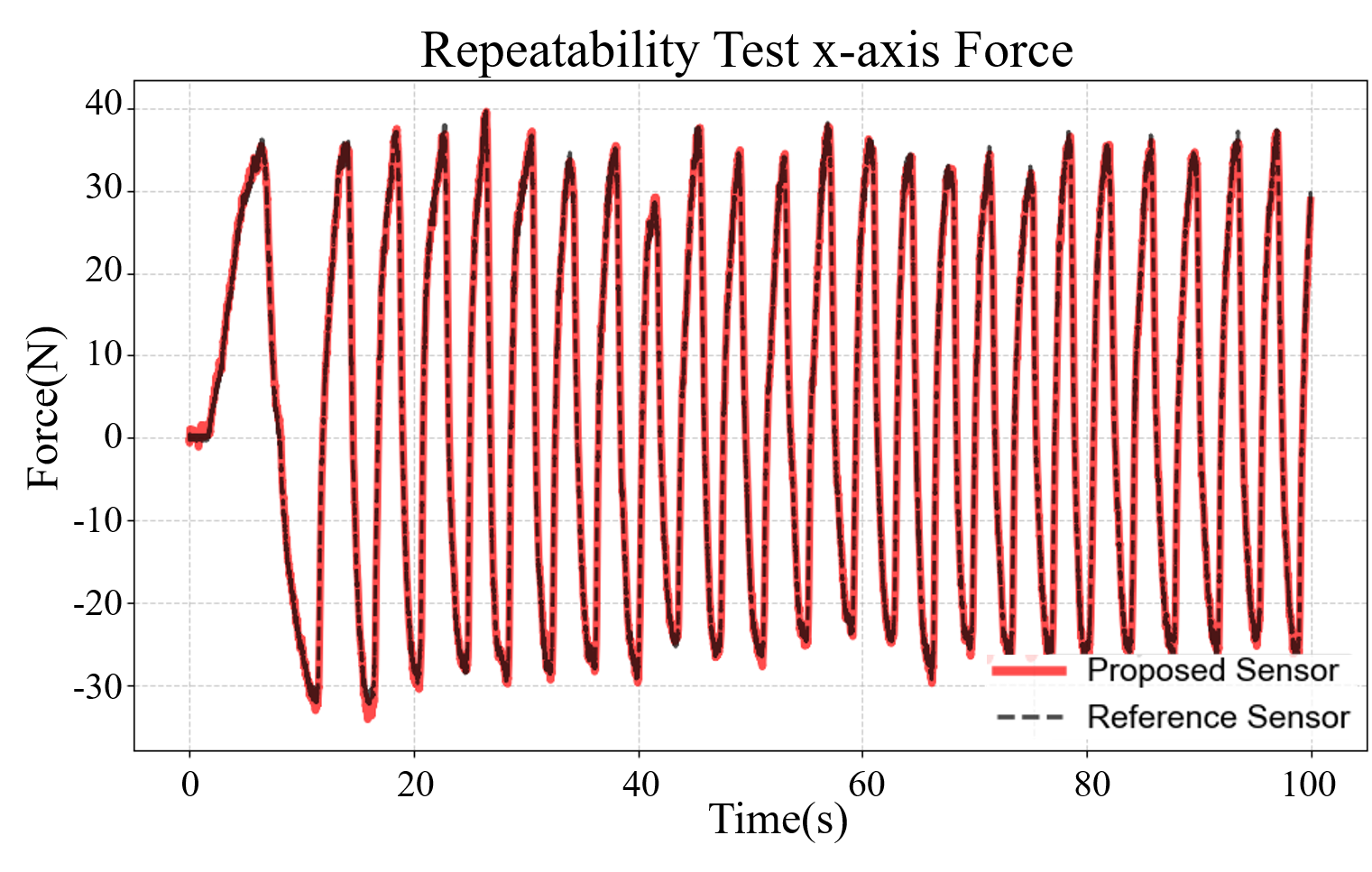}}
    \caption[Repeatability Test Result]{x-axis Force Repeatability Test Result during 100s.
    } \label{calrep}
\end{figure}

\begin{table}[!h]
\centering
\caption{Error Analysis, Nonlinearity, and Hysteresis}
\begin{tabular}{cccccc}
\hline
     & \multicolumn{3}{c}{Percentage Error (\%)} & \multicolumn{1}{c}{RMS Error} & \multicolumn{1}{c}{\makecell{Nonlinearity \\ Hysteresis }} \\ 
     & \multicolumn{1}{c}{Mean}    & \multicolumn{1}{c}{Std}    & Max    & \multicolumn{1}{c}{(N, N$\cdot$m)} & \multicolumn{1}{c}{(\%)} \\ \hline\hline
$F_x$ & 0.0460 & 0.0742 & 0.3600 & 0.5703 & 0.607  \\ 
$F_y$ & -0.0343 & 0.0706 & 0.3008 & 0.4051 & 0.521  \\ 
$F_z$ & 0.0172 & 0.0416 & 0.1316 & 0.6757 & 0.240  \\ 
$M_x$ & 0.0627 & 0.2669 & 0.8802 & 0.0172 & 1.543 \\ 
$M_y$ & 0.0570 & 0.2410 & 0.8672 & 0.0155 & 1.699  \\ 
$M_z$ & 0.0161 & 0.0852 & 0.3879 & 0.0063 & 0.681 \\ \hline
\end{tabular}
\label{combined_error_analysis}
\end{table}
Repeatability and crosstalk tests were conducted. The repeatability test was performed as shown in Figure~\ref{calrep}, and as summarized in Table~\ref{combined_repeatability_crosstalk}, the maximum percentage error was found to be as low as 0.8\%. 

\begin{figure*}[h]
    \centerline{\includegraphics[width=2\columnwidth]{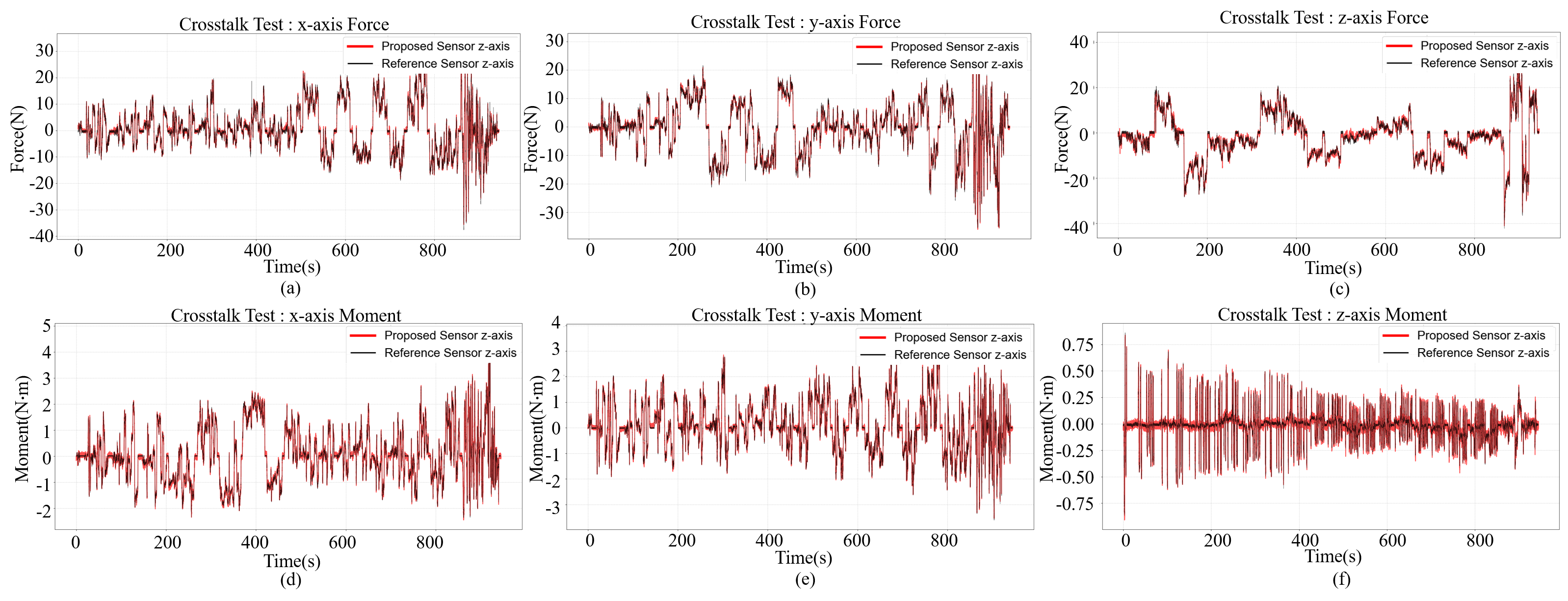}}
    \caption[Crosstalk Test]{Simultaneous Force and Moment Applying Experiment(Crosstalk Test Result) (a) Result of x-axis Force (b) Result of y-axis Force (c) Result of z-axis Force (d) Result of x-axis Moment (e) Result of y-axis Moment (f) Result of z-axis Moment 
    } \label{Crosstalk}
\end{figure*}

The crosstalk test was conducted as illustrated in Figure~\ref{Crosstalk}, where the maximum percentage error reached up to 3\% in the moment about the x-axis. However, the average RMSE was relatively small, suggesting that the error might have been caused by noise and the difference in rise time between the reference sensor and the tested system.

\begin{table}[!h]
\centering
\caption{Repeatability Test and Crosstalk Analysis}
\resizebox{\columnwidth}{!}{%
\begin{tabular}{ccccccc}
\hline
\multicolumn{7}{c}{\textbf{Repeatability Test}} \\ \hline\hline
& Mean & Std & Max & RMS Error (N) \\ 
$F_x$ & -0.0116 & 0.1759 & 0.8729 & 0.1432  \\ \hline 
\multicolumn{7}{c}{\textbf{Crosstalk Analysis}} \\\hline \hline 
& $F_x$ & $F_y$ & $F_z$ & $M_x$ & $M_y$ & $M_z$ \\ 
Percentage Error & 0.9422\% & 0.6310\% & 0.1827\% & 3.002\% & 1.974\% & 2.915\% \\ 
RMS Error & 0.7908N & 0.6460N & 1.132N & 0.05596N$\cdot$m & 0.05271N$\cdot$m & 0.01912N$\cdot$m \\ \hline
\end{tabular}%
}
\label{combined_repeatability_crosstalk}
\end{table}

A comparison was made with a commercially available sensor. The sensor used for comparison is the RFT40 from Robotus, which has a similar size of 40mm, an internal signal processing unit, and utilizes a non-contact capacitive type. The resolution comparison results are summarized in Table~\ref{resanal}. 

 \begin{figure*}[!h]
    \centerline{\includegraphics[width=1.8\columnwidth]{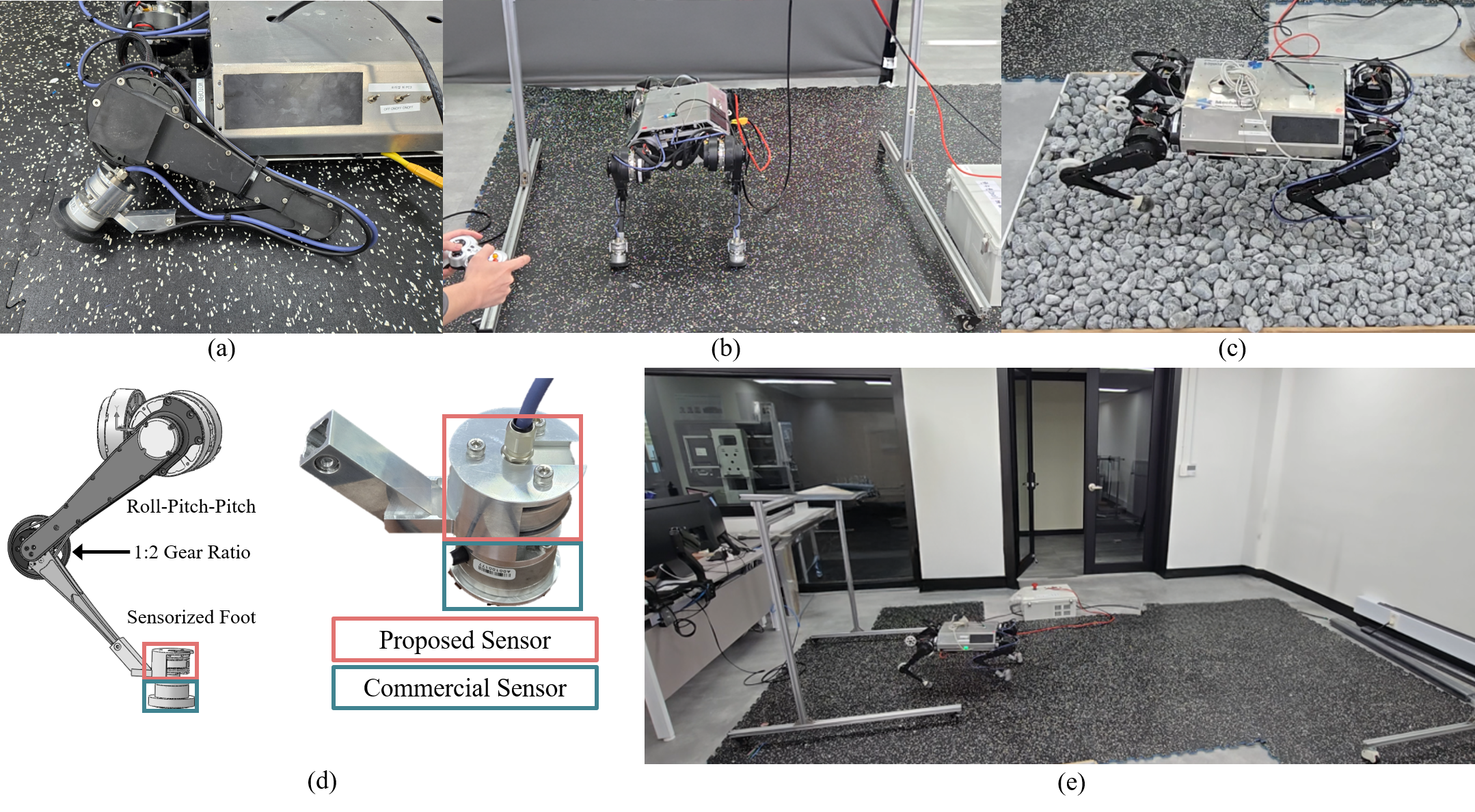}}
    \caption[Detailed Motion Experiment of Quadruped Robot]{Application of Proposed Sensor and Experiments for Verification of Proposed Sensor (a) A Picture of the foot with sensors (b) Posture Adjustment Experiment (c) Rough Terrain Experiment(Durability Test) (d) Attached Sensor Modules (e) Various Speed Experiment  } \label{detailmotion}
\end{figure*}

It can be observed that the proposed sensor demonstrates more than 10 times better resolution in the z-axis force and exhibits superior performance across all axes. Additionally, while the RFT40 has a sampling rate of 200Hz, the proposed sensor achieves a maximum sampling rate of 5kHz. Here, the term "steps" represents the number of distinguishable steps within the measurement range, calculated by dividing the measurement range by the resolution, which indicates the performance of the sensor in distinguishing different levels within its range.

\begin{table}[!h]
\centering
\caption{Resolution Analysis between Proposed Sensor and Commercial Sensor}
\resizebox{\columnwidth}{!}{
\begin{tabular}{ccccccc}
\hline
                & $F_x$      & $F_y$      & $F_z$      & $M_x$         & $M_y$        & $M_z$        \\ \hline\hline
RFT40           & 0.2N       & 0.2N       & 0.2N       & 8mNm          & 8mNm         & 8mNm         \\ 
Resolution(steps)      & 1000       & 1000       & 1500       & 625           & 625          & 625          \\ 
Proposed Sensor & 0.2436N    & 0.1429N    & 0.2017N    & 16mNm         & 21mNm        & 7.5mNm       \\ 
Resolution(steps)      & 5091       & 8258       & 19487      & 1716          & 1283         & 5272         \\ \hline
\end{tabular}
}
\label{resanal}
\end{table}

\section{Experiment of Quadruped Robot}
 \begin{figure*}[!ht]
    \centerline{\includegraphics[width=2\columnwidth]{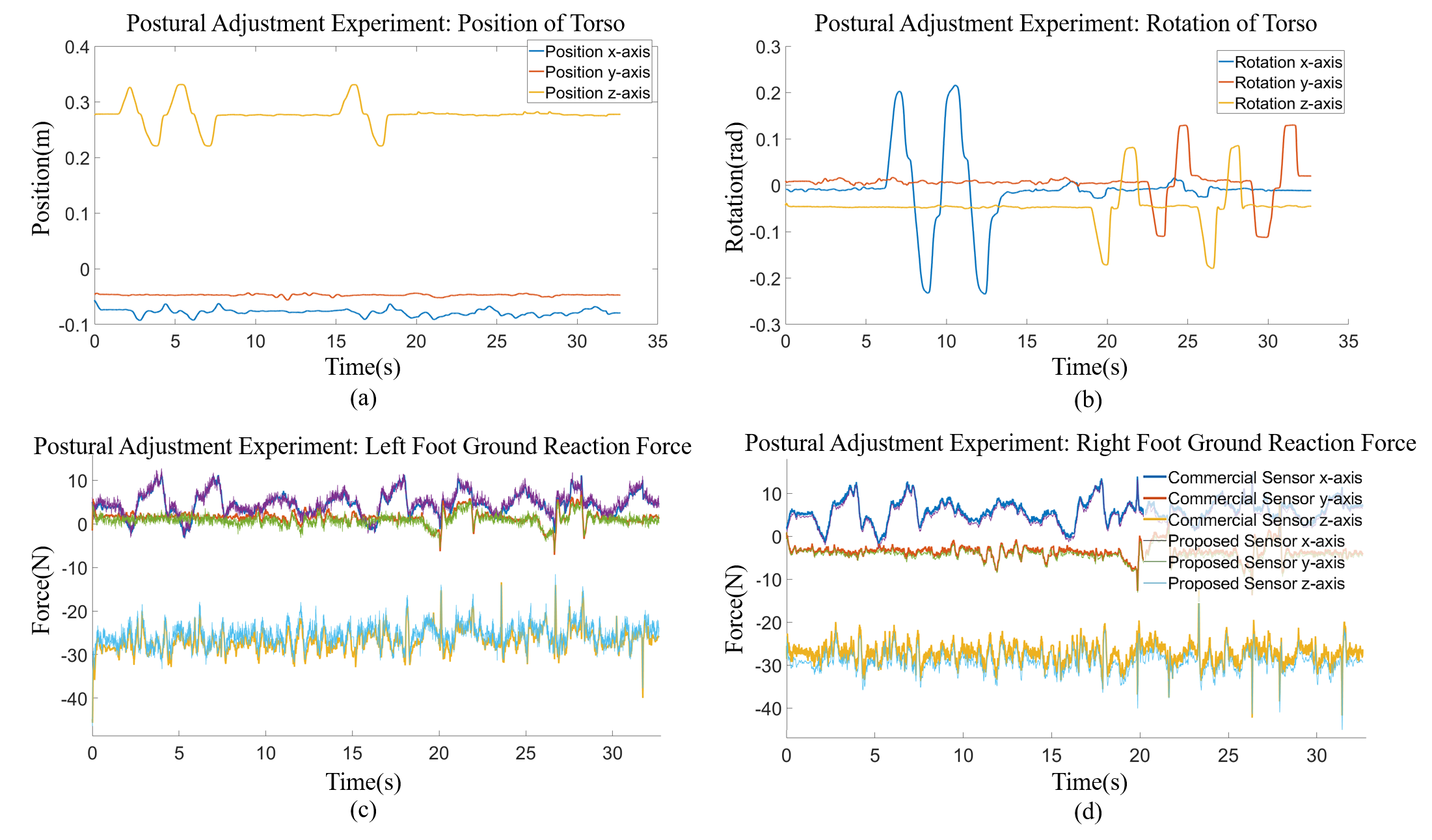}}
        \caption[Detailed Motion GRF Experiment of Quadruped Robot]{Postural Adjustment Experiment of A Quadruped Robot's Torso (a) Position Graph of Torso (b) Rotation Graph of Torso (c) Ground Reaction Force of Left Foot (d) Ground Reaction Force of Right Foot } \label{detailedmotiongrflf}
\end{figure*}
Unlike conventional manipulators, quadruped robots experience high-impact conditions. Typically, the impact force can reach 2 to 3 times the ground reaction force (GRF), which is directly related to the robustness of the sensor. Such repetitive impacts are a critical issue in quadruped robots. Therefore, experiments were conducted to evaluate whether the proposed sensor is suitable for quadruped robots. The objective was to verify its capability to measure small force variations, its robustness against high-impact conditions, and its performance across various speeds.


The proposed sensor was experimentally validated by mounting it on the foot of a quadruped robot. The RFT40 sensor was used as a reference sensor for comparison. Although the RFT40 has a smaller measurement range, its absolute accuracy is higher due to its narrower range, despite having the same relative accuracy of 1\%. The RFT40 is capable of measuring forces up to 150N in the z-axis and supports up to three times overload protection.

The quadruped robot used in the experiment was custom-built and has a maximum torque of 18N$\cdot$m. The robot weighs approximately 11kg, making it relatively lightweight. For the experiments, both the reference sensor (RFT40) and the proposed sensor were directly connected to the front legs of the robot. Due to the larger measurement range of the proposed sensor, it was placed on the upper side during the experiments.

The experiments consisted of three scenarios: posture adjustment, speed-based tests, and impact tests conducted on a gravel surface to simulate high-impact conditions.


Figure~\ref{detailmotion}(d) illustrates the sensor attachment on the foot, where the proposed sensor is positioned above and the commercial sensor is directly connected below. Both sensors communicate via CAN, as depicted in Figure~\ref{detailmotion}(a), which shows the signal transmission into the main body. Additionally, Figure~\ref{detailmotion}(b) presents an experiment in which postural variations are introduced to measure the ground reaction force, aiming to validate the proposed sensor by comparing it with the reference sensor. Figure~\ref{detailmotion}(c) demonstrates an experiment conducted on a gravel surface to evaluate the sensor's performance under high-impact conditions by comparing its measurements with those of the reference sensor. Lastly, Figure~\ref{detailmotion}(e) presents experimental results obtained at walking speeds of 0.2~m/s, 0.4~m/s, and 0.8~m/s.

The controller used in the experiment was a Customized Convex Model Predictive Controller along with a Whole-Body Controller, which were implemented based on~\cite{di2018dynamic, katz2019mini,kim2019highly}. For CAN communication, a Peak-CAN M.2 module was utilized. The control system was executed on a single-board computer, which simultaneously performed data logging and control tasks. The control signals were synchronized using Lightweight Communications and Marshalling (LCM)~\cite{huang2010lcm}. Additionally, the Chrono function was employed to log data with millisecond precision.

For the experiment, the RFT40 force-torque sensor from Robotus was used. The RFT40 is capable of measuring forces up to 150~N and has a maximum overload capacity of 450~N. The sensor was directly mounted for the experiment, and as a result, the forces along the x, y, and z axes, as well as the moment about the z-axis, could be expected to be equal. However, the moments about the x and y axes vary depending on the distance from the point where the force is applied. Therefore, only the forces along the x, y, and z axes were compared.

Typically, the elastomer of a force/torque sensor deforms by less than 0.1~mm, meaning that when directly connected, the same force can be assumed to be applied. However, in this setup, the RFT40 had its ground connected to the chassis. Although the robot is made of aluminum, its chassis was not grounded, leading to noise issues. To mitigate this, plastic insulation was used at the connection points when mounting the RFT40 to prevent electrical grounding interference.




\subsection{Postural adjustment experiment of a quadruped robot's torso}

First, an experiment was conducted to verify whether the proposed sensor can accurately measure subtle force variations while changing the posture of the quadruped robot. Since quadruped robots are floating-based systems, the ground reaction force at each foot plays a crucial role during posture adjustments. This experiment aimed to validate the sensor's capability in this context.

The experiment setup is illustrated in Figure~\ref{detailmotion}(b). During the experiment, the robot's torso was manipulated by altering its position along the z-axis and adjusting its roll, pitch, and yaw angles.

As shown in Table~\ref{combined_rms_max_error}, the RMSE was at most 1.97 N, and the maximum error was 0.36\%. The right foot exhibited slightly higher accuracy, which can be attributed to differences in sensor calibration and the fact that the robot's center of mass is not perfectly centered. Additionally, the hysteresis and deformation caused by inserting plastic insulation in the middle section when attaching the commercial sensor contributed to the observed errors.

\subsection{Speed adjustment experiment of a quadruped robot}
 \begin{figure*}[!ht]
    \centerline{\includegraphics[width=2\columnwidth]{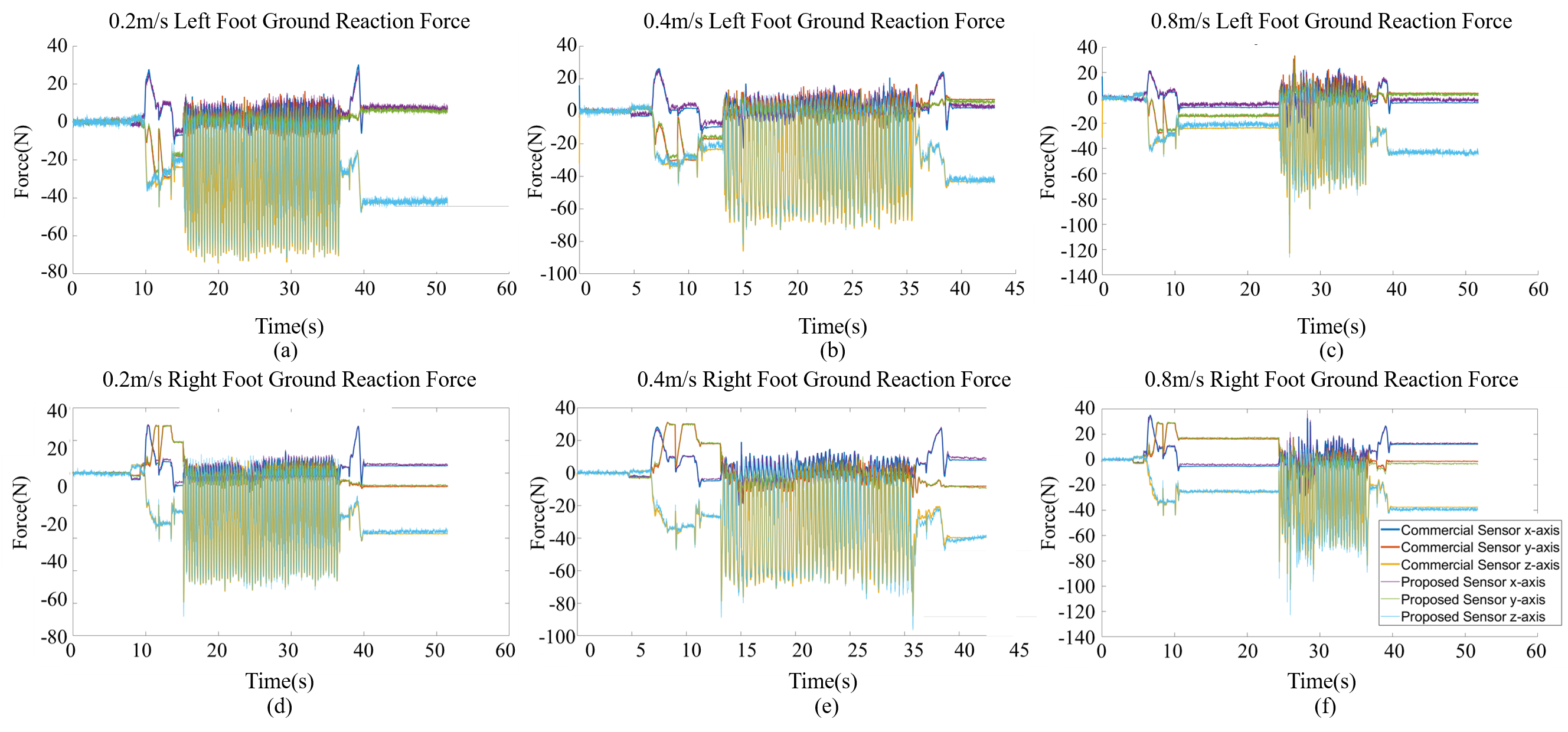}}
        \caption[Detailed Motion GRF Experiment of Quadruped Robot]{Speed Adjustment Experiment of A Quadruped Robot Result (a) Left Foot Ground Reaction Force at 0.2m/s (b) Left Foot Ground Reaction Force at 0.4m/s (c) Left Foot Ground Reaction Force at 0.8m/s (d) Right Foot Ground Reaction Force at 0.2m/s (e) Right Foot Ground Reaction Force at 0.4m/s (f) Right Foot Ground Reaction Force at 0.8m/s} \label{vasgrflf}
\end{figure*}
The next experiment involved measuring the ground reaction force (GRF) at speeds of 0.2 m/s, 0.4 m/s, and 0.8 m/s and comparing the results with the reference sensor. Since the force varies during acceleration from a stationary state to each speed, these transitions were analyzed. When the robot reaches its maximum speed, the ground reaction force generally stabilizes at similar values. However, at 0.8 m/s, maintaining balance becomes more challenging, requiring adjustments to roll and pitch balance, which results in increased force exertion. Although the sensor was directly attached, the deformation of the plastic insulation, previously mentioned as a structural component, contributed to measurement errors.

Figure~\ref{vasgrflf} illustrates the ground reaction forces for different speeds: (a), (b), and (c) correspond to the left foot at 0.2 m/s, 0.4 m/s, and 0.8 m/s, respectively, while (d), (e), and (f) represent the right foot under the same conditions. The results indicate that as speed increases, the force required to accelerate from rest to the target velocity also increases, leading to a larger ground reaction force. Furthermore, at 0.8 m/s, the quadruped robot's torso exhibits significant instability, necessitating greater ground reaction forces in the pitch and roll directions to maintain balance.

Each experiment began with the robot in a seated position, followed by a step input command for walking. In Figure~\ref{vasgrflf} (b), (c), (e), and (f), which correspond to the results for 0.4 m/s and 0.8 m/s, the force in the z-axis exhibits pronounced peaks during both acceleration at the beginning and deceleration toward the end. Table~\ref{combined_rms_max_error} shows that RMSE tends to increase with speed. Specifically, at 0.2 m/s, the RMSE reached a maximum of 3.96 N with a maximum percentage error of 1.70\%. At 0.4 m/s, the RMSE was at most 4.50 N, with a maximum percentage error of 1.72\%. At 0.8 m/s, the RMSE increased to 5.20 N, with a maximum percentage error of 1.97\%. 

As the applied force increased, deformation in the plastic insulation between the sensors became more pronounced. Additionally, since the proposed sensor has a sampling rate over five times higher than the reference sensor, impact forces resulted in greater discrepancies in force measurement. Nevertheless, the measured ground reaction forces remained within a range suitable for control applications.


\subsection{Rough Terrain Experiment of a Quadruped Robot}


The next experiment was conducted to verify whether the sensor remains functional under high-impact conditions by performing a walking test on a gravel surface. The gravel used in this experiment consisted of stones with diameters ranging from approximately 3 cm to 5 cm. Due to the robot’s active balance control, the ground reaction force increased. Additionally, as the sensor made direct contact with the gravel, impact forces were transmitted, making this experiment also a durability assessment.

\begin{figure}[!t]
    \centerline{\includegraphics[width=\columnwidth]{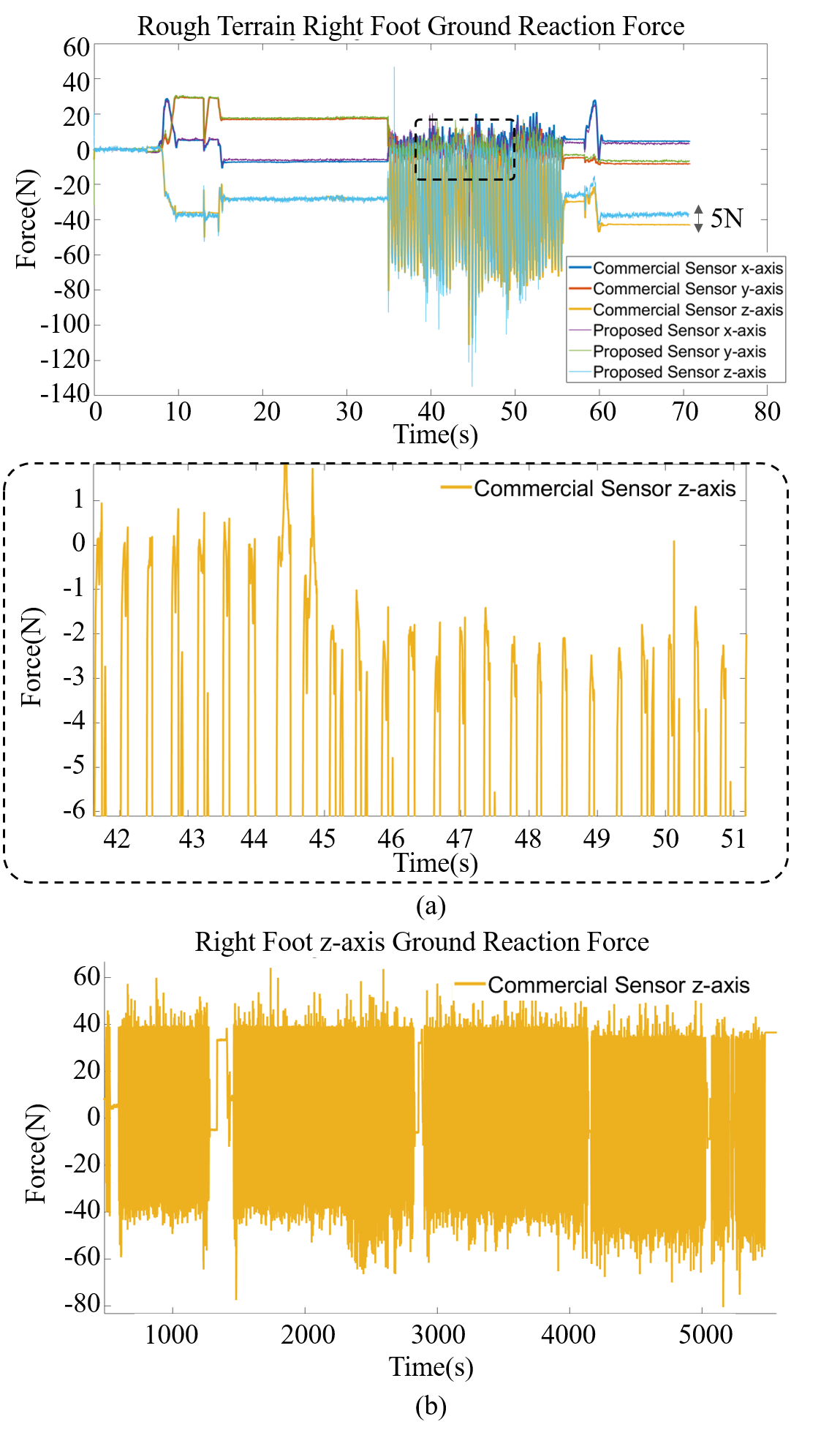}}
        \caption[Commercial Sensor's z-axis Force of Ground Reaction Force at Rough Terrain]{Rough Terrain Experiment Result(Durability Test) (a) Result of Left Foot Ground Reaction Force at Rough Terrain (b) Result of Right Foot Ground Reaction Force at Rough Terrain (c) Enlargement of (b) at 45s: Commercial Sensor Offset Occurs (d) Commercial Sensor's z-axis Right Foot Force of Ground Reaction Force at Rough Terrain} \label{roughrfz}
\end{figure}

The experiment followed the same procedure as the previous one, starting with the robot in a seated position, then standing up and walking toward the gravel surface. Over a duration of approximately 20 seconds, as shown in Figure~\ref{roughrfz}(a), a force of about 120 N was exerted, after which an offset appeared in the commercial sensor. Figure~\ref{roughrfz}(a) presents a magnified view of the force profile around the 40 to 50-second mark, showing that an impact event caused a 4 N force offset in the z-axis of the commercial sensor.

The z-axis force sensing capacity of the commercial sensor is 150 N, with an overload tolerance of 300\%, meaning it can withstand forces up to 450 N. However, in high-impact environments such as a gravel surface, transient forces can be several times greater than those encountered under normal ground reaction conditions. This can result in permanent deformation of the elastomer, affecting the sensor's performance.

To determine whether the offset observed on rough terrain was incidental, an extended walking experiment was conducted on the gravel surface for approximately 5000 seconds. Given that the quadruped robot takes approximately two steps per second, this corresponds to around 10,000 walking steps. The experiment was designed to continuously apply impact forces and observe any changes in the offset over time. This also serves as a durability test for the sensor.

As shown in Figure~\ref{roughrfz}(b), the results indicate that the z-axis readings of the commercial sensor exhibit a persistent offset throughout the experiment.

Figure~\ref{roughrfz}(b) illustrates the drift in the z-axis forces measured by the commercial sensors, which show a continuous change over time. For the right foot, the force steadily decreases, whereas for the left foot, the force alternates between decreasing and increasing repeatedly.

During the 5000-second experiment, the generalized coordinates were also logged. This experiment provided insights into the sensor's durability under such conditions.

\begin{table}[!h]
\centering
\caption{RMS Error and Max Percentage Error between Commercial Sensor and Proposed Sensor}
\resizebox{\columnwidth}{!}{%
\begin{tabular}{cccccccc}
\hline
\multicolumn{8}{c}{\textbf{Detailed Motion Experiment}} \\ \hline\hline
RMSE      & $F_x$(N)      & $F_y$(N)      & $F_z$(N)      & RMSE      & $F_x$(N)      & $F_y$(N)      & $F_z$(N)        \\ 
Left Foot    & 1.0845 & 1.0055 & 1.9748 & Right Foot   & 0.6997 & 0.7298 & 1.8449 \\ 
MAX Percentage Error      & $F_x$(\%)      & $F_y$(\%)      & $F_z$(\%)      & MAX Percentage Error       & $F_x$(\%)      & $F_y$(\%)      & $F_z$(\%)        \\ 
Left Foot & 0.3574 & 0.3331 & 0.1639 & Right Foot & 0.2009 & 0.2995 & 0.1792 \\ \hline
\multicolumn{8}{c}{\textbf{Various Speed Experiment (0.2 m/s)}} \\ \hline
RMSE      & $F_x$(N)      & $F_y$(N)      & $F_z$(N)      & RMSE      & $F_x$(N)      & $F_y$(N)      & $F_z$(N)        \\ 
Left Foot    & 2.0526 & 1.3997 & 3.9046 & Right Foot   & 1.7034 & 1.3007 & 3.9648 \\ 
MAX Percentage Error      & $F_x$(\%)      & $F_y$(\%)      & $F_z$(\%)      & MAX Percentage Error       & $F_x$(\%)      & $F_y$(\%)      & $F_z$(\%)        \\ 
Left Foot & 1.7015 & 1.1181 & 1.1140 & Right Foot & 1.0028 & 0.8712 & 0.7613 \\ \hline
\multicolumn{8}{c}{\textbf{Various Speed Experiment (0.4 m/s)}} \\ \hline
RMSE      & $F_x$(N)      & $F_y$(N)      & $F_z$(N)      & RMSE      & $F_x$(N)      & $F_y$(N)      & $F_z$(N)        \\ 
Left Foot    & 2.2480 & 1.4836 & 3.6591 & Right Foot   & 1.3974 & 1.5370 & 4.5013 \\ 
MAX Percentage Error      & $F_x$(\%)      & $F_y$(\%)      & $F_z$(\%)      & MAX Percentage Error       & $F_x$(\%)      & $F_y$(\%)      & $F_z$(\%)        \\ 
Left Foot & 1.7192 & 1.2257 & 0.9855 & Right Foot & 1.0305 & 0.8786 & 0.7790 \\ \hline
\multicolumn{8}{c}{\textbf{Various Speed Experiment (0.8 m/s)}} \\ \hline
RMSE      & $F_x$(N)      & $F_y$(N)      & $F_z$(N)      & RMSE      & $F_x$(N)      & $F_y$(N)      & $F_z$(N)        \\ 
Left Foot    & 2.7110 & 1.8895 & 4.2853 & Right Foot   & 2.1333 & 2.1510 & 5.1981 \\ 
MAX Percentage Error      & $F_x$(\%)      & $F_y$(\%)      & $F_z$(\%)      & MAX Percentage Error       & $F_x$(\%)      & $F_y$(\%)      & $F_z$(\%)        \\ 
Left Foot & 1.9716 & 1.4431 & 1.1507 & Right Foot & 1.2335 & 0.8817 & 0.9066 \\\hline
\multicolumn{8}{c}{\textbf{Rough Terrain Experiment}} \\ \hline
RMSE      & $F_x$(N)      & $F_y$(N)      & $F_z$(N)      & RMSE      & $F_x$(N)      & $F_y$(N)      & $F_z$(N)        \\ 
Left Foot    & 2.4558 & 2.2298 & 4.7692 & Right Foot   & 2.3660 & 2.4369 & 7.1704 \\ 
MAX Percentage Error      & $F_x$(\%)      & $F_y$(\%)      & $F_z$(\%)      & MAX Percentage Error       & $F_x$(\%)      & $F_y$(\%)      & $F_z$(\%)        \\ 
Left Foot & 1.8688 & 1.4903 & 1.3781 & Right Foot & 1.8518 & 1.7438 & 1.3683 \\ \hline
\end{tabular}%
}
\label{combined_rms_max_error}
\end{table}

As a result of the experiment, Table~\ref{Durabilitytestpercentage} compares the offsets of the commercial sensor and the proposed sensor for the left and right feet. The "before" values represent the offsets measured prior to the 5000-second experiment, while the "after" values indicate the offsets measured after the experiment. The commercial sensor exhibited a drift of up to 11.8 N in the z-axis. When the differences were expressed as a full-scale percentage error, the results are shown in Table~\ref{Durabilitytestpercentage}. The commercial sensor showed a drift of approximately 3.93\%, whereas the proposed sensor exhibited a maximum drift of 0.375\%. These results demonstrate that the proposed sensor has better durability compared to the commercial sensor, making it more suitable for application in quadruped robots.
\begin{table}[!h]
\centering
\caption{F/T Sensor Offset(Percentage) Before and After Durability Test}
\begin{tabular}{cccc}
\hline
       & $F_x$(\%)      & $F_y$(\%)      & $F_z$(\%)       \\ \hline\hline
Commercial Sensor LF & 2.27 &0.0300&3.93\\ 
Commercial Sensor RF & 0.270&0.670&3.18\\ 
Proposed Sensor LF & 0.375 &0.165&0.116\\ 
Proposed Sensor RF& 0.0381& 0.107&0.0354\\ \hline
\end{tabular}

\label{Durabilitytestpercentage}
\end{table}

\section{Conclusion and Future works}

The objective of this study was to develop a robust 6-axis force-torque sensor for quadruped robots. The sensor was designed to be compact while capable of measuring large forces, with an optimized structure to enhance resolution. A novel approach was proposed using photo-couplers to implement the entire sensing mechanism on a single PCB, resulting in a compact 6-axis force-torque sensor. Additionally, a method was developed to optimize the sensor's positioning and elastomer parameters through precise modeling, improving both numerical accuracy and sensitivity. The proposed sensor was integrated into a quadruped robot for experimental validation.

Compared to commercial sensors, the proposed sensor requires fewer components, is more affordable, and can measure a larger force range. It demonstrated an improvement in resolution by a factor of at least 2, and up to 10 times in some cases. Unlike traditional strain gauge-based sensors, which typically require external signal processing units due to their size limitations, this sensor allows for integrated signal processing within a single device by utilizing photo-couplers.

Additionally, experiments were conducted by applying the proposed sensor to a quadruped robot and comparing it with a commercial sensor. The results demonstrated that the proposed sensor outperformed the commercial sensor in terms of durability. The contactless commercial sensor exhibited a maximum drift of 3.93\%, while the proposed sensor showed only 0.375\%. On average, the commercial sensor displayed an offset of 1.725\%, whereas the proposed sensor had an average offset of 0.1394\%, which is more than 10 times smaller. These results highlight the durability and reliability of the proposed sensor compared to the commercial alternative.

While photo-couplers, as analog sensors, have limitations, such as the potential need for additional analog filters and circuitry to achieve higher resolution, they offer advantages by reducing the number of components compared to traditional strain gauge systems, thereby lowering the likelihood of failure. These attributes make the sensor highly suitable for compact applications, offering significant potential for the robotics industry.

\newpage

\section{Biography Section}
\vspace{11pt}
\begin{IEEEbiography}[{\includegraphics[width=1in,height=1.25in,clip,keepaspectratio]{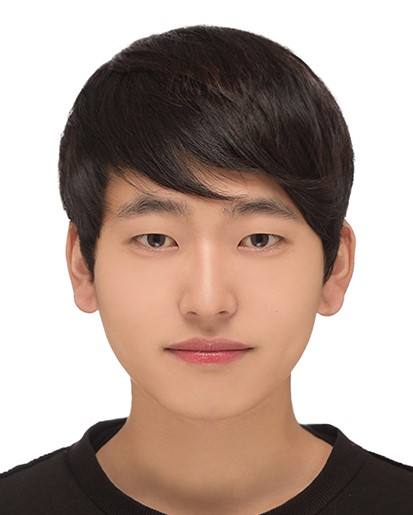}}]{Hyun-Bin Kim}
~ received the B.S., M.S. and Ph.D. degrees in mechanical engineering from Korea Advanced Institute of Science and Technology(KAIST), Daejeon, Republic of Korea, in 2020, 2022 and 2025 respectively. He is currently working as the post-doctor researcher in KAIST. His current research interests include force/torque sensors, legged robot control, robot design and mechatronics system.
\end{IEEEbiography}
\vspace{11pt}
\begin{IEEEbiography}[{\includegraphics[width=1in,height=1.25in,clip,keepaspectratio]{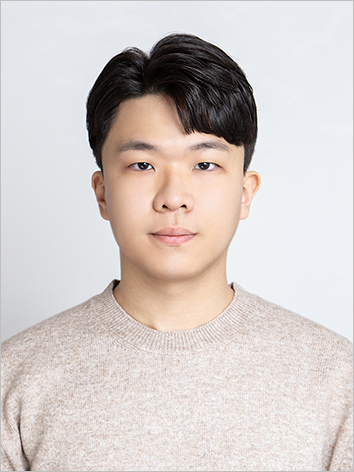}}]{Byeong-Il Ham}
~ received the B.S. degree in school of robotics from University of Kwangwoon, Seoul, and the M.S. degree in robotics program from Korea Advanced Institute of Science and Technology(KAIST), Daejeon, Republic of Korea, in 2022 and 2024, respectively. He is in Doctor Program in KAIST, Daejeon, Korea, from 2024. His current research interests include legged system, optimal control and motion planning.
\end{IEEEbiography}
\vspace{11pt}
\begin{IEEEbiography}[{\includegraphics[width=1in,height=1.25in,clip,keepaspectratio]{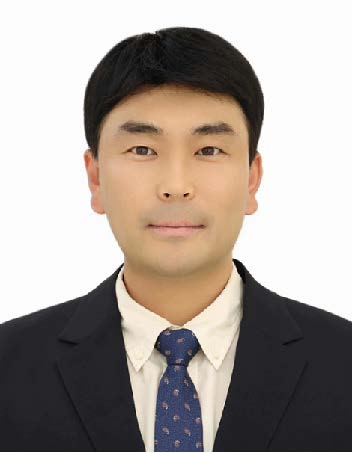}}]{Keun-Ha Choi}
~ received the B.S. degree in weapon system engineering from Korea Military Academy, Seoul, and the M.S. degree and the Ph.D. degrees in mechanical engineering from Korea Advanced Institute of Science and Technology(KAIST), Daejeon, Republic of Korea in 2002, 2007 and 2016 respectively. He was a Defense Acquisition Program Administration, a Project Management Officer, Republic of Korea, from 2009 to 2012 and from 2016 to 2019, Army Education and Doctrine Command, an AI Research Officer, from 2019 to 2020, and an Army Headquarters, Force Planning Officer, from 2020 to 2021. In 2021, he was an AI/Big Data Research Officer with the Army Future Innovation Research Center. In 2022, he joined as a Research Assistant Professor with Daedong-KAIST Research Center for Mobility. Since 2023, he has been with the Department of Mechanical Engineering, KAIST. His current research interests include sensor fusion-based robot autonomous navigation algorithm and control, vision sensor-based object detection using AI, and defense AI application plan/military operation concept/concept design.
\end{IEEEbiography}
\vspace{11pt}
\begin{IEEEbiography}[{\includegraphics[width=1in,height=1.25in,clip,keepaspectratio]{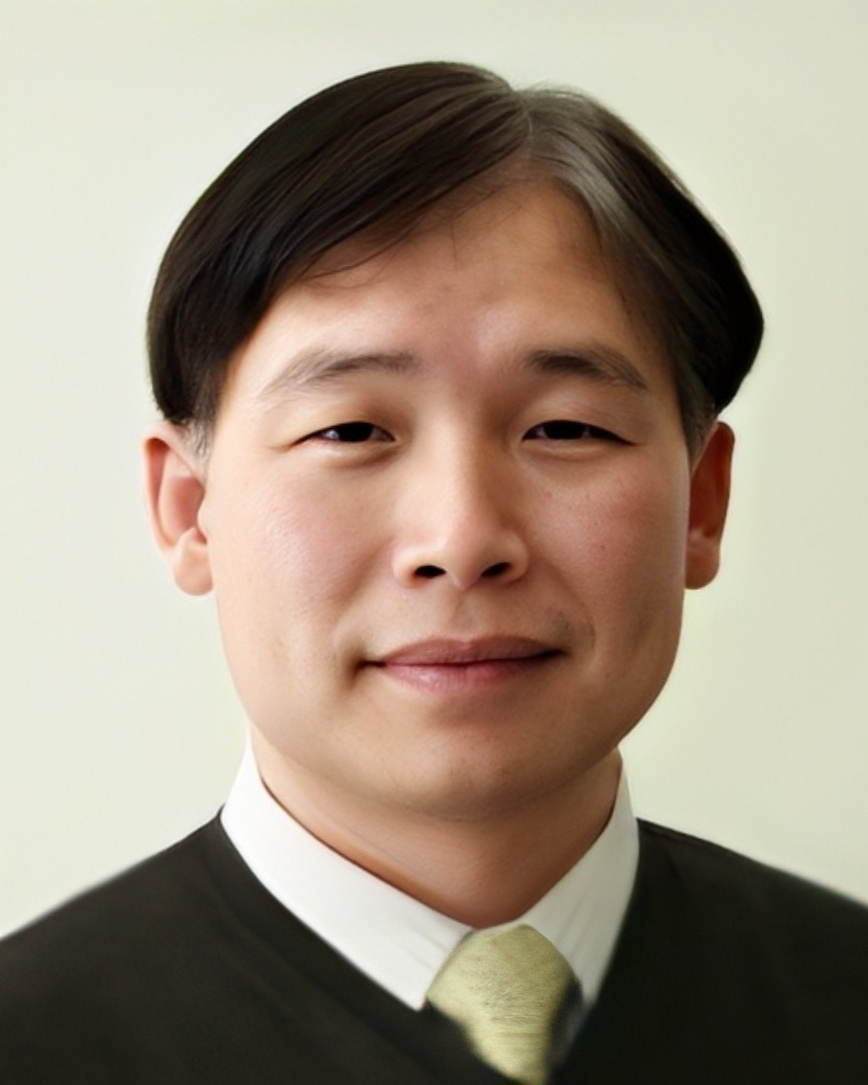}}]{Kyung-Soo Kim}~(Fellow, IEEE)~ received the B.S., M.S., and Ph.D. degrees in mechanical engineering from Korea Advanced Institute of Science and Technology (KAIST), Daejeon, Republic of Korea, in 1993, 1995, and 1999, respectively. He was a Chief Researcher with LG Electronics Inc., from 1999 to 2003, and the DVD Group Manager of STMicroelectronics Company Ltd., from 2003 to 2005. In 2005, he joined the Department of Mechanical Engineering, Korea Polytechnic University, Siheung, Republic of Korea, as a Faculty Member. Since 2007, he has been with the Department of Mechanical Engineering, KAIST. His research interests include control theory, electric vehicles, and autonomous vehicles. He serves as an Associate Editor for the Automatica and the Journal of Mechanical Science and Technology.
\end{IEEEbiography}

\end{document}